\documentclass[10pt,twocolumn,letterpaper]{article}

\usepackage{iccv}
\usepackage{times}
\usepackage{epsfig}

\usepackage{graphicx}
\usepackage{amsmath}
\usepackage{amssymb}
\usepackage{booktabs}
\usepackage{multirow}
\usepackage{color}
\usepackage{capt-of}
\usepackage{cuted}
\usepackage{xfrac}
\usepackage{pifont}
\newcommand{\cmark}{\ding{51}}%
\newcommand{\tabincell}[2]{\begin{tabular}{@{}#1@{}}#2\end{tabular}}

\usepackage{xcolor}
\definecolor{citecolor}{HTML}{0071BC}
\definecolor{linkcolor}{HTML}{ED1C24}

\usepackage[breaklinks=true,bookmarks=false]{hyperref}

\usepackage[capitalize]{cleveref}
\crefname{section}{Sec.}{Secs.}
\Crefname{section}{Section}{Sections}
\Crefname{table}{Table}{Tables}
\crefname{table}{Tab.}{Tabs.}

\iccvfinalcopy 

\def\iccvPaperID{****} 

\ificcvfinal\fi

\begin{document}

\title{Image Inpainting via Iteratively Decoupled Probabilistic Modeling}

\author{%
	Wenbo Li\textsuperscript{1} \quad Xin Yu\textsuperscript{2} \quad Kun Zhou\textsuperscript{3} \quad Yibing Song\textsuperscript{4} \quad Zhe Lin\textsuperscript{5} \quad Jiaya Jia\textsuperscript{1} \vspace{0.1in} \\ 
	$^{1}$CUHK  \quad  ${^2}$The University of Hong Kong \quad
	${^3}$CUHK (Shenzhen) \quad ${^4}$Tencent AI Lab \quad ${^5}$Adobe Inc. \\
	{\tt\small \{wenboli,leojia\}@cse.cuhk.edu.hk \quad yuxin27g@gmail.com} \\
	{\tt\small kunzhou@link.cuhk.edu.cn \quad yibingsong.cv@gmail.com \quad zlin@adobe.com}
	\vspace{-0.45in}
}

\maketitle
\ificcvfinal\thispagestyle{empty}\fi

\begin{strip}\centering
	\includegraphics[width=1.0\linewidth]{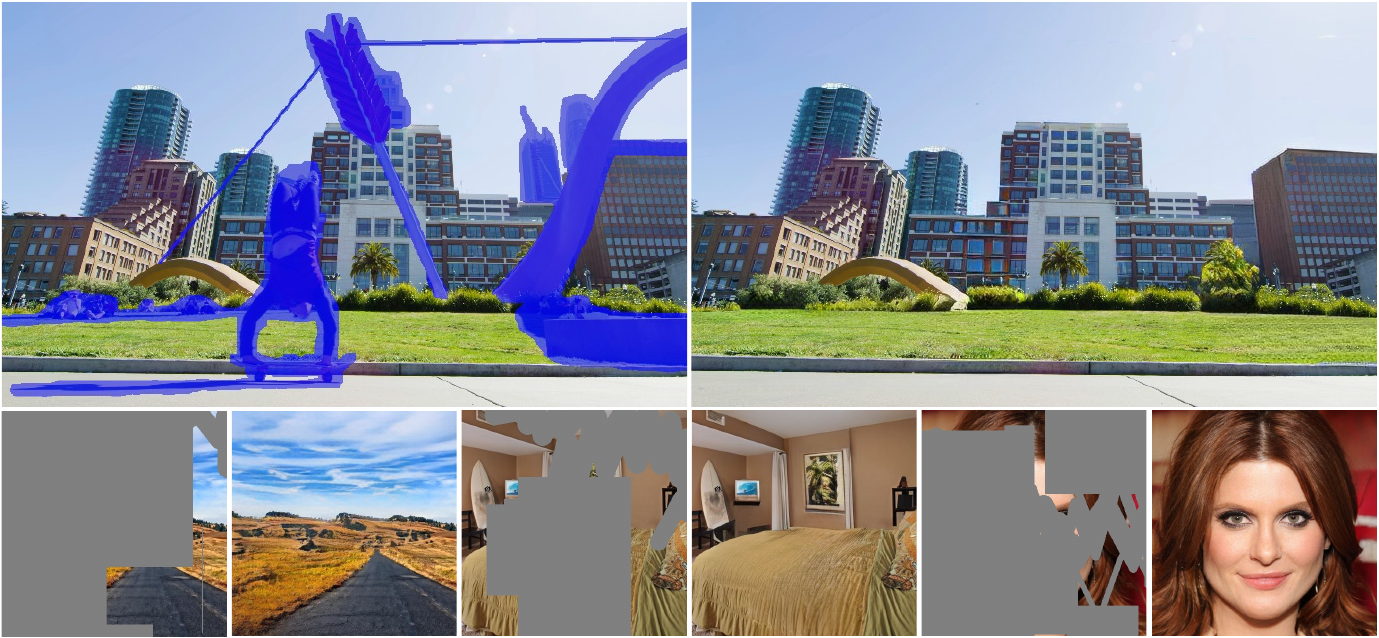}
	\captionof{figure}{	Our model supports photo-realistic large-hole inpainting for various scenarios. The first example for object removal is a high-resolution image captured in the wild, while other inpainting examples ($512 \times 512$) come from Places2~\cite{zhou2017places} and CelebA-HQ~\cite{karras2018progressive} datasets. \label{fig:teasing}}
\end{strip}

\begin{abstract}
	Generative adversarial networks (GANs) have made great success in image inpainting yet still have difficulties tackling large missing regions. In contrast, iterative probabilistic algorithms, such as autoregressive and denoising diffusion models, have to be deployed with massive computing resources for decent effect. To achieve high-quality results with low computational cost, we present a novel pixel spread model (PSM) that iteratively employs decoupled probabilistic modeling, combining the optimization efficiency of GANs with the prediction tractability of probabilistic models. As a result, our model selectively spreads informative pixels throughout the image in a few iterations, largely enhancing the completion quality and efficiency. On multiple benchmarks, we achieve new state-of-the-art performance. Code is released at \url{https://github.com/fenglinglwb/PSM}. \\ \vspace{-0.53in}
	
	%
\end{abstract}

\section{Introduction}
\label{sec:intro}

Image inpainting, a fundamental computer vision task, aims to fill the missing regions in an image with visually pleasing and semantically appropriate content. It has been extensively employed in graphics and imaging applications, such as photo restoration~\cite{wan2020bringing,wan2022old}, image editing~\cite{barnes2009patchmatch,jo2019sc}, compositing~\cite{levin2004seamless}, re-targeting~\cite{cho2017weakly}, and object removal~\cite{criminisi2004region}. This task, especially filling large holes, is more ill-posed than other restoration problems, necessitating models of stronger generation abilities.



In past years, generative adversarial networks (GANs) have made great processes in image inpainting~\cite{pathak2016context,yan2018shift,yu2018generative,liu2019coherent,wan2021high,li2022mat}. By implicitly modeling a target distribution through a min-max game, GANs-based methods significantly outperform traditional exemplar-based techniques~\cite{hays2007scene,sun2005image,criminisi2004region,criminisi2003object} in terms of visual quality. However, the one-shot generation of GANs sometimes lead to unstable training~\cite{salimans2016improved,gulrajani2017improved,kodali2017convergence} and makes it challenging to learn a complex distribution, particularly when inpainting large holes in high-resolution images.

Conversely, autoregressive models~\cite{van2016conditional,van2016pixel,parmar2018image} and denoising diffusion models~\cite{song2019generative,ho2020denoising,dhariwal2021diffusion} recently demonstrated remarkable power in content generation~\cite{ramesh2022hierarchical,saharia2022photorealistic,yu2022scaling,singer2022make}. These models utilize tractable probabilistic modeling techniques to iteratively refine the image based on prior estimations, resulting in more stable training and improved coverage. However, it is widely known that autoregressive models process images pixel by pixel, which makes it cumbersome to handle high-resolution data. On the other hand, denoising diffusion models typically require thousands of iterations to achieve accurate estimations. Thus, using these methods directly in image inpainting incurs respective drawbacks -- {\it strategies for high-quality large-hole high-resolution image inpainting still fall short}.

\begin{figure}[t]
	\begin{center}
		\includegraphics[width=1.0\linewidth]{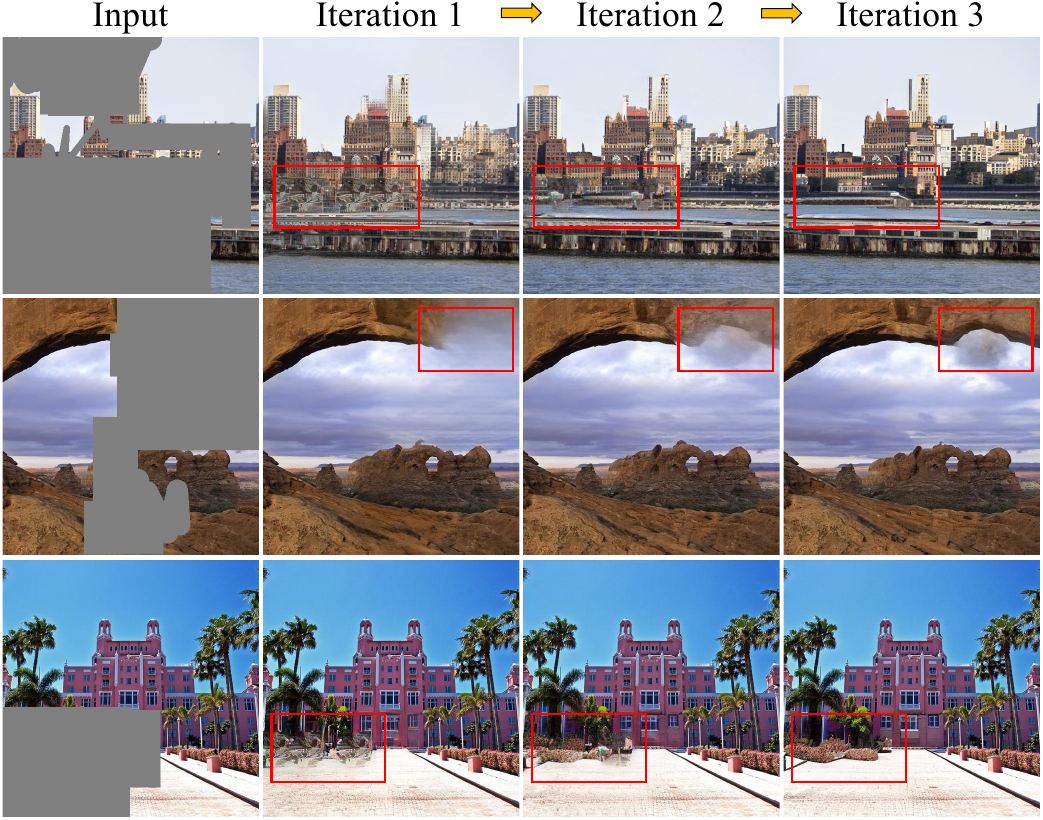}
	\end{center}
	\vspace{-0.1in}
	\caption{Inpainting results of our PSM at different iterations. One-shot generation usually results in blurry content with unpleasing artifacts, while more iterations yield better results.}
	\label{fig:prog}
	\vspace{-0.15in}
\end{figure}

To complete the map of inpainting, in this paper, we develop a new pixel spread model (PSM) tailored for the large-hole scenario. PSM operates in an iterative manner, where all pixels are predicted in parallel during each iteration, and only qualified predictions are retained for subsequent iterations. It acts as a process to gradually spread trustful pixels to unknown locations. Our core design lies in a simple yet highly effective decoupled probabilistic modeling (see \cref{sec:decouple}), which enjoys the merits of GANs' efficient optimization and the tractability of probabilistic models. In detail, our model simultaneously predicts an inpainted result (\ie, the mean term) and an uncertainty map (\ie, the variance term). The mean term is optimized using implicit adversarial training, yielding more accurate predictions with fewer iterations. The variance term, contrarily, is modeled explicitly using Gaussian regularization. 

The adoption of our decoupled strategy offers numerous advantages. First, the use of adversarial optimization leads to a significant reduction in the number of iterative steps required to achieve promising results, as shown in Fig.~\ref{fig:prog}, much faster than autoregressive and denoising diffusion models. Second, the Gaussian regularization employed produces a variance term that naturally acts as an uncertainty measure (see \cref{sec:spa}). This allows for the selection of reliable estimates for iterative refinement, largely reducing GANs' artifacts. Furthermore, the explicit modeling of the distribution facilitates continuous sampling, thereby producing predictions with enhanced quality and diversity, as demonstrated in \cref{sec:exp}. Finally, the uncertainty measure is instrumental in constructing an uncertainty-guided attention mechanism (see \cref{sec:arch}), which encourages the network to leverage more informative pixels for efficient reasoning. As a result, our PSM completes large missing regions with photo-realistic content, as illustrated in Fig.~\ref{fig:teasing}.

Our contributions can be summarized as follows:
\begin{itemize}
	\vspace{-0.1in}
	\item We develop a novel pixel spread model (PSM) customized for large-hole image inpainting. Thanks to the proposed iteratively decoupled probabilistic modeling, our model achieves efficient optimization and high-quality completion.
	\vspace{-0.1in}
	\item Our method reaches cutting-edge performance on both Places~\cite{zhou2017places}
	and CelebA-HQ~\cite{karras2018progressive} benchmark datasets. Notably, our PSM outperforms popular denoising diffusion models, \eg, LDM~\cite{rombach2022high}, by a large margin, yielding 1.1 FID improvement on Places2~\cite{zhou2017places} while being significantly more light-weighted (only 20\% parameters, $10\times$ faster).
\end{itemize}

\section{Related Work}
\label{sec:rela}

\subsection{Traditional Methods}
Image inpainting is a classical computer vision problem. Early methods make use of image priors, such as self-similarity and sparsity. Diffusion-based methods~\cite{bertalmio2000image, ballester2001filling}, for instance, convey information to the holes from nearby undamaged neighbors. Another line of exemplar-based approaches~\cite{hays2007scene,sun2005image,le2011examplar,criminisi2003object,ding2018image,lee2016laplacian} looks for highly similar patches to complete missing regions using human-defined distance metrics. The most representative work is PatchMatch~\cite{barnes2009patchmatch}, which employs heuristic searching in a multi-scale image space to speed up inpainting greatly. However, due to a lack of context understanding, they do not guarantee visually appealing and semantically consistent results.

\begin{figure*}[t]
	\begin{center}
		\includegraphics[width=1.0\linewidth]{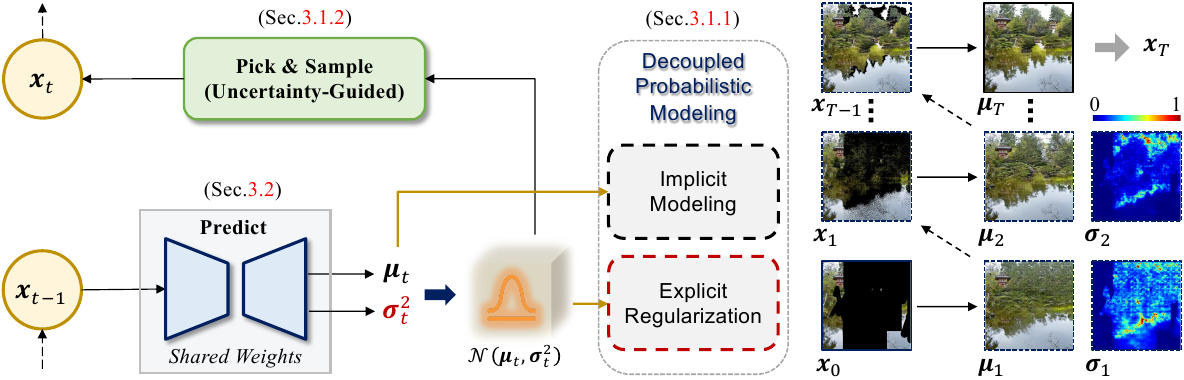}
	\end{center}
	\vspace{-0.05in}
	\caption{Our pixel spread model for high-quality large-hole image inpainting. Left illustration is the pixel spread pipeline with proposed decoupled probabilistic modeling, and the right images are visual examples. We simplify the input of the $t$-th iteration to $\boldsymbol{x}_{t-1}$, and denote the estimated mean and variance as $\boldsymbol{\mu}_t$ and $\boldsymbol{\sigma}^2_t$. The $\boldsymbol{\sigma}_t$ map on the right is normalized for better visualization. We observe gradual uncertainty reduction in missing regions during the pixel spread process.}
	\vspace{-0.15in}
	\label{fig:framework}
\end{figure*}

\subsection{Deep Learning Based Methods}
Using a great amount of training data to considerably increase the ability of high-level understanding, deep-neural-network-based methods~\cite{pathak2016context,yan2018shift,zeng2019learning,liu2020rethinking,wang2018image} achieve success. Pathak \etal~\cite{pathak2016context} introduce the adversarial loss~\cite{goodfellow2014generative} to inpainting, yielding visually realistic results. Several approaches along this line continually push the performance to new heights. For example, in order to obtain locally fine-grained details and globally consistent structures, Iizuka \etal~\cite{iizuka2017globally} adopt two discriminators for adversarial training. Additionally, partial~\cite{liu2018image} and gated~\cite{yu2019free} convolution layers are proposed to reduce artifacts, \eg, color discrepancy and blurriness, for irregular masks. Moreover, intermediate cues, including foreground contours~\cite{xiong2019foreground}, object structures~\cite{nazeri2019edgeconnect,ren2019structureflow}, and segmentation maps~\cite{song2018spg} are used in multi-stage generation. Despite nice inpainting content for small masks, these methods still do not guarantee large-hole inpainting quality. 

\subsection{Large Hole Image Inpainting}
To deal with large missing regions, a surge of effort was made to improve the model capability. Attention techniques~\cite{yu2018generative,liu2019coherent,xie2019image,yi2020contextual} and transformer architectures~\cite{wan2021high,zheng2021tfill,li2022mat} take advantage of context information. They work well when an image contains repeating patterns. Besides, Zhao \etal~\cite{zhao2020large} propose a novel architecture, bridging the gap between image-conditional and unconditional generation, improving free-form large-scale image completion. There are also attempts to study the progressive generation. This line is to select only high-quality pixels each time and gradually fill holes. We note that these methods heavily rely on specially designed update algorithms~\cite{zhang2018semantic,guo2019progressive,li2020recurrent,oh2019onion}, or consume additional model capacity to separately assess the prediction accuracy~\cite{zeng2020high}, or need more training stages~\cite{chang2022maskgit} when processing images. 

Recently, benefiting from exact likelihood computation and iterative samplings, autoregressive models~\cite{wan2021high,yu2021diverse,wu2022nuwa} and denoising diffusion models~\cite{saharia2022palette,rombach2022high,lugmayr2022repaint,avrahami2022blended} have shown great potential in producing diversified and realistic content. They inevitably incur high inference costs with thousands of steps and require massive computation resources. In this work, we present decoupled probabilistic modeling that obtains predictions and uncertainty measures simultaneously. Our model identifies reliable predicted pixels and sends them to subsequent iterations, thereby mitigating GANs-generated artifacts. Also, the proposed approach can be viewed as a diffusion model that learns pixel spreading rather than denoising and requires fewer iterations.


\section{Our Method}
\label{sec:method}

Our objective is to use photo-realistic material to complete a masked image with substantial missing areas. In this section, we first formulate our pixel spread model (PSM) along with a comprehensive analysis. It is followed by the details of model design and loss functions.

\subsection{Pixel Spread Model}
\label{sec:form}

Although GANs-based methods achieve significantly better results than traditional ones, they still face great difficulties handling large missing regions. We attribute one of the reasons to the one-shot nature of GANs and instead propose iterative inpainting. 

In each pass, since there are inevitably some good predictions, we use these pixels as clues to assist the next-time generation. In this way, our pixel spread model gradually propagates valuable information to the entire image. In the following, we first discuss the single-pass modeling before moving on to the pixel spread process.

\subsubsection{Decoupled Probabilistic Modeling}
\label{sec:decouple}

For iterative inpainting, it is essential to find a mechanism to evaluate the accuracy of predictions. One intuitive solution is introducing a tractable probabilistic model so that uncertainty information can be analytically calculated. However, this requirement often leads to the assumption that the approximated target distribution is Gaussian, which is considerably too simple to explain the truly complicated distributions. Although several iterative models, such as denoising diffusion models~\cite{ho2020denoising}, enrich the expression of marginal distribution by including a number of hidden variables and optimizing the variational lower evidence bound, these methods typically yield a high inference cost.

To address these key issues, we propose a decoupled probabilistic modeling tailored for efficient iterative inpainting. The essential insight is that we leverage the advantages of implicit GANs-based optimization and explicit Gaussian regularization {\it in a decoupled way}. Thus we can simultaneously obtain accurate predictions and explicit uncertainty measures. 

As shown in \cref{fig:framework}, given an input image $\boldsymbol{x}_t$ at time $t$ with large holes, our model (see architecture details in Sec.~\ref{sec:arch}) predicts the inpainting result $\boldsymbol{\mu}_t$ as well as an uncertainty map $\boldsymbol{\sigma}_t^2$. We use the adversarial loss (along with other losses of Sec.~\ref{sec:loss}) to supervise image prediction $\boldsymbol{\mu}_t$, while jointly treating $(\boldsymbol{\mu}_t,\boldsymbol{\sigma}_t^2)$ as the mean and diagonal covariance of Gaussian distribution. GANs' implicit optimization makes it possible to approximate the true distribution as closely as possible, greatly reducing the number of iterations. It also supplies us with an explicit uncertainty measure for the mean term, allowing us to select reliable pixels. The Gaussian regularization is mainly applied to the variance term using negative log likelihood (NLL) $\mathcal{L}_{nll}$ as
\begin{equation}
\small
\mathcal{L}_{nll} = -\sum_{i=1}^{D} \log \int_{\delta_{-}(\boldsymbol{y}^{i})}^{{\delta_{+}(\boldsymbol{y}^{i})}} \mathcal N \left(y; {\rm sg} [\boldsymbol{\mu}^{i}_{\theta}(\boldsymbol{x})], {\boldsymbol{\sigma}^{i}_{\theta}}(\boldsymbol{x})^2 \right) dy \,,
\end{equation}
where $D$ is the data dimension and $i$ is the pixel index, $\theta$ denotes model parameters, and input $\boldsymbol{x}$ and output $\boldsymbol{y}$ are scaled to $[-1, 1]$. $\delta_{+}(y)$ and $\delta_{-}(y)$ are defined as
\begin{align}
\delta_{+}(y) &=\left\{
\begin{array}{cl}
\infty           & {\rm if} \ y \ = \ 1 \,, \\
y+\frac{1}{255} & {\rm if} \ y \ < \ 1 \,,
\end{array}
\right. \\
\delta_{-}(y) &=\left\{
\begin{array}{cl}
-\infty           & {\rm if} \ y \ = \ -1 \,, \\
y-\frac{1}{255} & {\rm if} \ y \ > \ -1 \,.
\end{array}
\right.
\end{align}
Specifically, we include a stop-gradient operation (\ie, ${\rm sg[\cdot]}$), which encourages the Gaussian constraint only to optimize the variance term and allows the mean term to be estimated from more accurate implicit modeling.

\vspace{0.05in}
\noindent\textbf{Discussion.} We use the estimated mean and variance for sampling during the diffusion process, while taking the deterministic mean term as the output for the final iteration. The feasibility of this design is proved by the experiments in Sec.~\ref{sec:exp}. Additionally, the probabilistic modeling enables us to apply continuous sampling during pixel spread, yielding higher quality and more diverse estimations. Finally, we find the uncertainty measure also enables us to design a more effective attention mechanism in Sec.~\ref{sec:arch}.

\subsubsection{Pixel Spread Scheme}
\label{sec:spa}

We use a feed-forward network, denoted as $f_{\theta}(\cdot)$, to gradually spread informative pixels to the entire image, starting from known regions as
\begin{equation}
\label{eq:iter}
\boldsymbol{x}_{t}, \boldsymbol{m}_{t}, \boldsymbol{u}_{t} = f_{\theta}(\boldsymbol{x}_{t-1}, \boldsymbol{m}_{t-1}, \boldsymbol{u}_{t-1}) \,,
\end{equation}
where $t$ is the time step, $\boldsymbol{x}_{t}$ refers to the masked image, $\boldsymbol{m}_{t}$ stands for a binary mask, and $\boldsymbol{u}_{t}$ is the uncertainty map. The output includes the updated image, mask, and uncertainty map. Network parameters are shared across all iterations.

We use several iterations for both training and testing to improve performance. Specifically, as shown in Fig.~\ref{fig:framework} and Eq.~\eqref{eq:iter}, our method runs as follows at the $t$-th iteration.
\begin{enumerate}
	\item \textbf{Predict.} Given the masked image $\boldsymbol{x}_{t-1}$, mask $\boldsymbol{m}_{t-1}$, and uncertainty map $\boldsymbol{u}_{t-1}$, our method estimates mean $\boldsymbol \mu_t$ and variance $\boldsymbol \sigma^2_t$ for all pixels.  Then a preliminary uncertainty map $\tilde{\boldsymbol{u}}_{t}$ scaled to $\left[0,1\right]$ is generated by converting the variance map.
	
	\item \textbf{Pick.} We first sort the uncertainty scores for unknown pixels. According to the pre-defined mask schedule, we calculate the number of pixels that are to be added in this iteration, and insert those with the lowest uncertainty to the known category, updating the mask to $\boldsymbol{m}_{t}$. Based on the preliminary uncertainty map $\tilde{\boldsymbol{u}}_{t}$, by marking locations that are still missing as 1 and the initially known pixels as 0, we obtain the final uncertainty map $\boldsymbol{u}_{t}$.
	
	\item \textbf{Sample.} We consider two situations. First, for the initially known locations $\boldsymbol{m}_{0}$, we always use the original input pixels $\boldsymbol{x}_{0}$. Second, we apply continuous sampling in accordance with $\boldsymbol \mu_t$ and $\boldsymbol \sigma_t$ for the inpainting areas. The result is formulated as
	\begin{equation}
	\label{eq:update}
	\boldsymbol{x}_{t} = \boldsymbol{x}_{0} + (\boldsymbol{m}_{t} - \boldsymbol{m}_{0}) \odot (\boldsymbol{\mu_t} + \alpha \cdot \boldsymbol{\sigma}_t \odot \boldsymbol{z}) \,,
	\end{equation}
	where $\alpha$ is an adjustable ratio and $\boldsymbol{z} \sim \mathcal N (\boldsymbol{0}, \boldsymbol{I})$, and $\odot$ denotes the Hadamard product. Note that we do not use the $\boldsymbol{\sigma}_t \boldsymbol{z}$ term in the final iteration.
\end{enumerate}

\subsection{Model Architecture}
\label{sec:arch}

We use a deep U-Net~\cite{ronneberger2015u} architecture with a StyleGAN~\cite{karras2019style,karras2020analyzing} decoder, reaching large receptive fields with stacked convolutions to take advantage of context information in images~\cite{buades2005non,mairal2009non,berman2016non,wang2018non}. In addition, we introduce multiple attention blocks at various resolutions, in light of the discovery that global interaction significantly improves reconstruction quality on much larger and more diverse datasets at higher resolutions~\cite{yu2018generative,yi2020contextual,dhariwal2021diffusion}.

\begin{figure}[t]
	\begin{center}
		\includegraphics[width=1.0\linewidth]{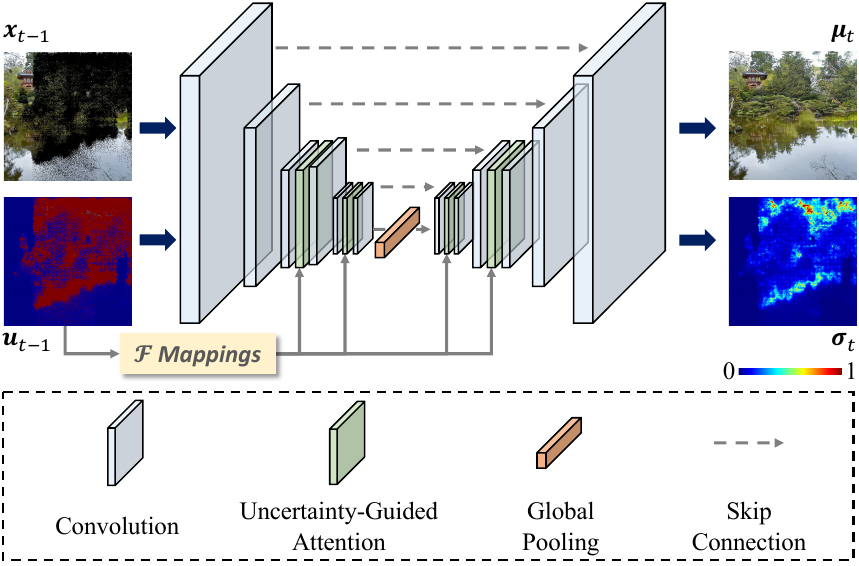}
	\end{center}
	\vspace{-0.1in}
	\caption{U-Net architecture with the proposed uncertainty-guided attention. We omit the mask update for clarity.}
	\label{fig:unet}
\end{figure}

Based only on feature similarity, the conventional attention mechanism~\cite{vaswani2017attention} offers equal opportunity for pixels to exchange information. For the inpainting task, however, missing pixels are initialized with the same specified values, making them close to one another. As a result, it is usually unable to effectively leverage useful information from visible regions. Even worse, the valid pixels are compromised, resulting in blurry content and unpleasing artifacts. 

In this situation, as shown in Fig.~\ref{fig:unet}, we take into account the pixels' uncertainty scores to adjust the aggregating weights in attention. It properly resolves the problem mentioned above. The attention output is computed by
\begin{equation}
{\rm Attention}(\boldsymbol{q}, \boldsymbol{k}, \boldsymbol{v}, \boldsymbol{u})={\rm Softmax} \left(\frac{\boldsymbol{q}\boldsymbol{k}^T}{\sqrt{d_k}} + \mathcal{F}(\boldsymbol{u}) \right) \boldsymbol{v} \,,
\end{equation}
where $\{\boldsymbol{q}, \boldsymbol{k}, \boldsymbol{v}\}$ are the query, key, value matrices, $d_k$ denotes the scaling factor, and $\mathcal{F}$ represents learnable functions that predict biased pixel weights based on the uncertainty map $\boldsymbol{u}$ and also include a reshape operation.

\subsection{Loss Functions}
\label{sec:loss}
In each iteration, our model outputs the mean and variance estimates, as shown in Fig~\ref{fig:framework}. The mean term is optimized using adversarial loss~\cite{goodfellow2014generative} $\mathcal{L}_{adv}$ and perceptual loss~\cite{suvorov2021resolution,johnson2016perceptual} $\mathcal{L}_{pcp}$, which aims to produce natural-looking images. The losses are described as follows.

\vspace{0.05in}
\noindent\textbf{Adversarial loss.} We formulate the adversarial loss as
\begin{align}
\mathcal{L}_{ag} & = - \mathbb{E}_{\hat x} \left[ \log \left(D \left( \hat x \right) \right) \right] \,, \\
\mathcal{L}_{ad} & = - \mathbb{E}_{x} \left[ \log \left( D \left( x \right) \right) \right] - \mathbb{E}_{\hat x} \left[ \log \left( 1 - D \left( \hat x \right) \right) \right] \,,
\end{align}
where $D$ is the discriminator implemented as~\cite{karras2019style}, $x$ and $\hat x$ are real and predicted images.

\vspace{0.05in}
\noindent{\textbf{Perceptual loss.}}
We adopt a high receptive filed perceptual loss~\cite{suvorov2021resolution}, which is formulated as
\begin{equation}
\mathcal{L}_{pcp} = \sum_{i}\left\| \phi_i(x) - \phi_i(\hat x) \right\|^2_2 \,,
\end{equation}
where $\phi_{i}$ is the layer output of a pre-trained ResNet50~\cite{he2016deep}.

As discussed in Sec.~\ref{sec:decouple}, we apply the negative log likelihood $\mathcal{L}_{nll}$ to constrain the variance for uncertainty modeling. Thus the final loss function for the generator is
\begin{equation}
\mathcal{L} = \sum\nolimits_{j} \lambda_{1}  \mathcal{L}_{ag}^{j} + \lambda_{2} \mathcal{L}_{pcp}^{j}  + \lambda_{3}  \mathcal{L}_{nll}^{j},
\end{equation}
where $j$ is the number of spread iterations. We empirically set $\lambda_{1}=1$, $\lambda_{2}=2$ and $\lambda_{3}$ to $1 \times 10^{-4}$.

\section{Experiments}
\label{sec:exp}

\subsection{Datasets and Metrics}

We train our models at $512 \times 512$ resolution on Places2~\cite{zhou2017places} and CelebA-HQ~\cite{karras2018progressive} in order to adequately assess the proposed method. Places2 is a large-scale dataset with nearly 8 million training images in various scene categories. Additionally, 36,500 images make up the validation split. During training, images undergo random flipping, cropping, and padding, while testing images are centrally cropped to the $512 \times 512$ size. For CelebA-HQ, we employ 24,183 and 2,993 images, respectively, to train and test our models. Following~\cite{yu2019free,zhao2020large,suvorov2021resolution,li2022mat}, we use on-the-fly generated masks during training, where the detailed setup is from MAT~\cite{li2022mat}. We evaluate all models using identical masks provided by~\cite{li2022mat} for fair comparisons. Besides, for evaluating model robustness, we use the same model to inpaint both small and large masks.

Despite being adopted in early inpainting work, L1 distance, PSNR, and SSIM~\cite{wang2004image} are found not strongly associated with human perception when assessing image quality~\cite{ledig2017photo,sajjadi2017enhancenet}. In this work, in light of~\cite{zhao2020large,li2022mat}, we use FID~\cite{heusel2017gans}, P-IDS~\cite{zhao2020large}, and U-IDS~\cite{zhang2018unreasonable}, which robustly measures the perceptual fidelity of inpainted images, as more suitable metrics.

\subsection{Implementation Details}

We use an encoder-decoder architecture. The encoder is made up of convolution blocks, while the decoder is adopted from StyleGAN2~\cite{karras2020analyzing}. The encoder's channel size starts at 64 and doubles after each downsampling until the maximum of 512. The decoder has a symmetrical configuration. We adopt attention blocks at $32 \times 32$ and $16 \times 16$ resolutions for both the encoder and decoder. The uncertainty map is initialized as ``1 - mask'' at the first iteration. Given an $H \times W$ input, we first downsample the feature size to $\frac{H}{32} \times \frac{W}{32}$ before returning to $H \times W$. 

We train our models for 20M images on Places2 and CelebA-HQ using 8 NVIDIA A100 GPUs. We utilize exponential moving average (EMA), adaptive discriminator augmentation (ADA), and weight modulation training strategies~\cite{karras2020training,li2022mat}. The batch size is 32, and the learning rate is $1\times10^{-3}$. We employ an Adam~\cite{kingma2015adam} optimizer with $\beta_{1} = 0$ and $\beta_{2} = 0.99$. We empirically set $\alpha=0.01$ in Eq.~\eqref{eq:update} based on experimental results. To generate $512 \times 512$ images, we iterate twice for training and four times for testing.

The fact that previous work~\cite{zhao2020large,li2022mat} trains models on Places2 with 50M or more images -- much more extensive data than ours -- evidences the benefit of our method. It is found that additional training can further improve our approach, and yet 20M images currently already deliver cutting-edge performance.

\subsection{Ablation Study}

\begin{table}[t]
	\renewcommand\arraystretch{1.2}
	\begin{center}
		\resizebox{0.95\linewidth}{!}{
			\begin{tabular}{c | c c c c c | c}
				\hline
				Model & Iter. & DPM & CS & UGA & Att. Res. & FID$\downarrow$ \\
				\hline
				A & 3 & \cmark & \cmark & \cmark & 32,16 & \textbf{2.36} \\
				\hline
				B & 1 &     -      &      -    & \cmark & 32,16  & 2.95 \\
				C & 2 & \cmark & \cmark & \cmark & 32,16 & 2.55 \\
				D & 3 &            & \cmark & \cmark & 32,16  & 2.49\\
				E & 3 & \cmark &            & \cmark & 32,16  & 2.45 \\
				F & 3 & \cmark & \cmark &            & 32,16  & 2.44 \\
				G & 3 & \cmark & \cmark & \cmark &    16   &  2.64\\
				\hline
			\end{tabular}	
		}
	\end{center}
	\vspace{-0.05in}
	\caption{Quantitative ablation study of the number of training iterations, modeling and sampling strategies, and architecture designs. Model ``A'' refers to the full model. Models ``B'' and ``C'' use fewer iterations. We tease apart the decoupled probabilistic modeling (DPM), continuous sampling (CS), and uncertainty-guided attention (UGA) in models ``D'', ``E'', and ``F''. Model ``G'' only adopts attention blocks at $16 \times 16$ resolution.}
	\label{tab:ablation}
\end{table}

\begin{table}[t]
	\renewcommand\arraystretch{1.15}
	\begin{center}
		\resizebox{0.99\linewidth}{!}{
			\begin{tabular}{c | c c c c c c c c}
				\hline
				Iter. & 4 & 5 & 6 & 7 & 8 & 9 & 10 \\
				\hline
				Model A & 2.23 & 2.16 & 2.12 & 2.09 & 2.07 & \textbf{2.05} & \textbf{2.05} \\
				Model B & 2.27 & 2.20 & 2.16 & 2.14 & 2.12 & 2.11 & 2.11 \\
				\hline
			\end{tabular}	
		}
	\end{center}
	\vspace{-0.05in}
	\caption{Quantitative ablation study of the number of testing iterations. As the number of iterations increases, the FID$\downarrow$ results get better and then saturate.}
	\label{tab:test_iter}
\end{table}

In this section, we conduct a comprehensive investigation of the proposed designs in our method. For quick evaluation, we train our models for 6M images at $256 \times 256$ resolution using Places365-Standard, a subset of Places2~\cite{zhou2017places}.  We start with model ``A'', which employs our full designs and adopts three iterations during training.

\begin{figure}[t]
	\begin{center}
		\includegraphics[width=1.0\linewidth]{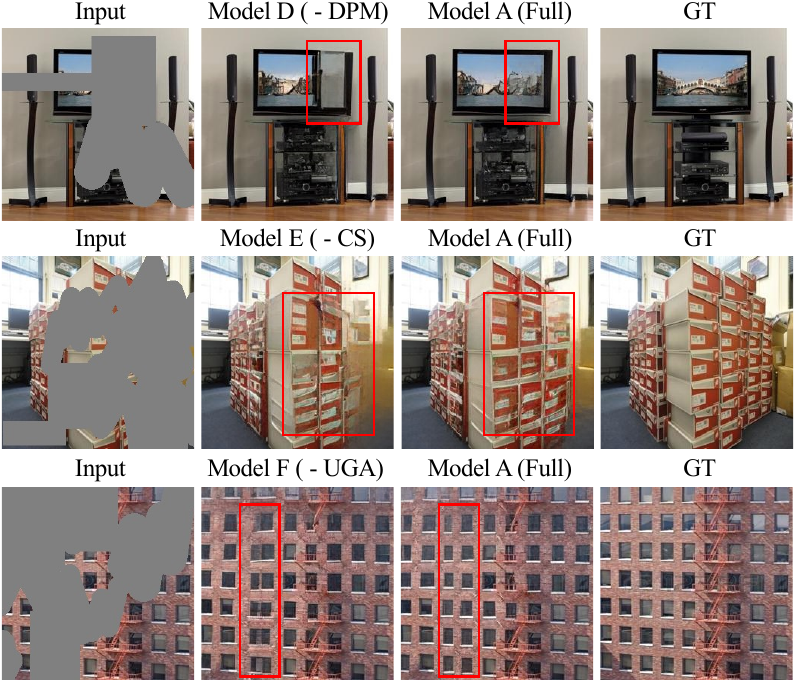}
	\end{center}
	\vspace{-0.05in}
	\caption{Qualitative ablation study. Model ``A'' is our full model. The proposed decoupled probabilistic modeling, continuous sampling, and uncertainty-guided attention designs are not used in models ``D'', ``E'', and ``F''.}
	\label{fig:abl_vis}
\end{figure}

\vspace{0.05in}
\noindent\textbf{Iterative number.} Our core idea is to employ iterative optimization to enhance the generation quality. We adjust the iteration number and maintain the same setup during training and testing. As illustrated in Table~\ref{tab:ablation}, models with one and two iterations, dubbed ``B" and ``C", yield 0.59 and 0.19 FID decreases compared to model ``A''. Additionally, as shown in Fig.~\ref{fig:prog}, adopting more iterations is capable of producing more aesthetically pleasing content. The first and third cases exhibit obviously fewer artifacts, and the arch in the second example is successfully restored after three iterations. Both the quantitative and qualitative results manifest the importance of iterative generation.

\begin{table*}[t]
	\renewcommand\arraystretch{1.25}
	\setlength\tabcolsep{2pt}
	\begin{center}
		\resizebox{\linewidth}{!}{
			\begin{tabular}{c | c | c c c | c c c | c c c | c c c }
				\hline
				\multirow{3}{*}{Method} & \multirow{3}{*}{\tabincell{c}{\#Param. \\ $\times 10^6$}} & \multicolumn{6}{|c}{Places2 ($512 \times 512$)} & \multicolumn{6}{|c}{CelebA-HQ ($512 \times 512$)} \\
				\cline{3-14}
				~ & ~ & \multicolumn{3}{|c|}{Small Mask} & \multicolumn{3}{|c|}{Large Mask} & \multicolumn{3}{|c|}{Small Mask} & \multicolumn{3}{|c}{Large Mask} \\
				\cline{3-14}
				~ & ~ & FID$\downarrow$ & P-IDS($\%$)$\uparrow$ & U-IDS($\%$)$\uparrow$ & FID$\downarrow$& P-IDS($\%$)$\uparrow$ & U-IDS($\%$)$\uparrow$ & FID$\downarrow$ & P-IDS($\%$)$\uparrow$ & U-IDS($\%$)$\uparrow$ & FID$\downarrow$ & P-IDS($\%$)$\uparrow$ & U-IDS($\%$)$\uparrow$ \\
				\hline
				PSM (ours) & 74 & \textcolor{red}{0.72} & \textcolor{red}{30.95} & \textcolor{red}{43.91} & \textcolor{red}{1.68} & \textcolor{red}{25.33} & \textcolor{red}{39.30} & \textcolor{red}{2.34} & \textcolor{red}{22.42} & \textcolor{red}{33.43} & \textcolor{red}{4.05} & \textcolor{red}{16.10} & \textcolor{red}{28.25} \\
				\hline
				Stable Diffusion$^\dagger$ & 860 & 1.32 & 12.69 & 34.78 & \textcolor{blue}{2.11} & 12.01 & 32.57 & - & - & - & - & - & - \\
				LDM~\cite{rombach2022high} & 387 & \textcolor{blue}{1.06} & 16.23 & 39.61 & 2.76 & 12.11 & 33.02 & - & - & - & - & - & - \\
				MAT~\cite{li2022mat} & 62 & 1.07 & \textcolor{blue}{27.42} & \textcolor{blue}{41.93} & 2.90 & 19.03 & 35.36 & \textcolor{blue}{2.86} & \textcolor{blue}{21.15} & \textcolor{blue}{32.56} & \textcolor{blue}{4.86} & \textcolor{blue}{13.83} & \textcolor{blue}{25.33} \\
				CoModGAN~\cite{zhao2020large} & 109 & 1.10 & 26.95 & 41.88 & 2.92 & \textcolor{blue}{19.64} & \textcolor{blue}{35.78} & 3.26 & 19.65 & 31.41 & 5.65 & 11.23 & 22.54 \\
				LaMa~\cite{suvorov2021resolution} & 51/27 & 0.99 & 22.79 & 40.58 & 2.97 & 13.09 & 32.29 & 4.05 & 9.72 & 21.57 & 8.15 & 2.07 & 7.58 \\
				ICT~\cite{wan2021high} & 150 & - & - & - & - & - & - & 6.28 & 2.24 & 9.99 & 12.84 & 0.13 & 0.58 \\
				MADF~\cite{zhu2021image} & 85 & 2.24 & 14.85 & 35.03 & 7.53 & 6.00 & 23.78 & 3.39 & 12.06 & 24.61 & 6.83 & 3.41 & 11.26 \\
				AOT GAN~\cite{zeng2021aggregated} & 15 & 3.19 & 8.07 & 30.94 & 10.64 & 3.07 & 19.92 & 4.65 & 7.92 & 20.45 & 10.82 & 1.94 & 6.97 \\	
				HFill~\cite{yi2020contextual} & 3 & 7.94 & 3.98 & 23.60 & 28.92 & 1.24 & 11.24 & - & - & - & - & - & - \\
				DeepFill v2~\cite{yu2019free} & 4 & 3.02 & 9.17 & 32.56 & 9.27 & 4.01 & 21.32 & 10.11 & 3.11 & 9.52 & 24.42 & 0.17 & 0.42\\
				EdgeConnect~\cite{nazeri2019edgeconnect} & 22 & 4.03 & 5.88 & 27.56 & 12.66 & 1.93 & 15.87 & 10.58 & 4.14 & 12.45 & 39.99 & 0.10 & 0.22 \\
				\hline  
			\end{tabular}
		}
	\end{center}
	\vspace{-0.05in}
	\caption{Quantitative comparisons on Places~\cite{zhou2017places} and CelebA-HQ~\cite{karras2018progressive}. ``$\dagger$'': the officially released Stable Diffusion inpainting model trained on a large-scale high-quality dataset LAION-Aesthetics V2 5+. Our method achieves the best performance under both large and small mask settings. The \textcolor{red}{best} and \textcolor{blue}{second best} results are in red and blue.} 
	\label{tab:sota}
\end{table*}

It is noted that we can test the system with a different number of iterations from the training stage. Using more iterations results in higher FID performance, as demonstrated in Table~\ref{tab:test_iter}, yet at the expense of longer inference time. Thus, there is a trade-off between inference speed and generation quality. Additionally, when comparing models ``A'' and ``B'', it is clear that introducing more iterations in the training process is beneficial. But the number of iterations in the inference stage is more important.

\vspace{0.05in}
\noindent\textbf{Decoupled probabilistic modeling.} To deliver accurate prediction while supporting the uncertainty measure for iterative inpainting, we propose decoupled probabilistic modeling. When putting all supervision on the sampled result, we observe the training diminishes the variance term (close to 0 for all pixels). It is because, unlike denoising diffusion models that precisely quantify the noise levels at each step, our GANs-based method no longer provides specific optimization targets for the mean and variance terms. The variance term is underestimated for trivial optimization in this case. It renders the picking process less effective. 

As illustrated in Table~\ref{tab:ablation}, model ``D'' obtains an inferior FID result compared with the full model ``A''. Besides, from the visual comparison in Fig.~\ref{fig:abl_vis}, it is observed that model ``D'' tends to generate blurry content, while model ``A'' produces sharper structures and fine-grained details. All these observations prove the effectiveness of this design.

\begin{figure}[t]
	\begin{center}
		\includegraphics[width=1.0\linewidth]{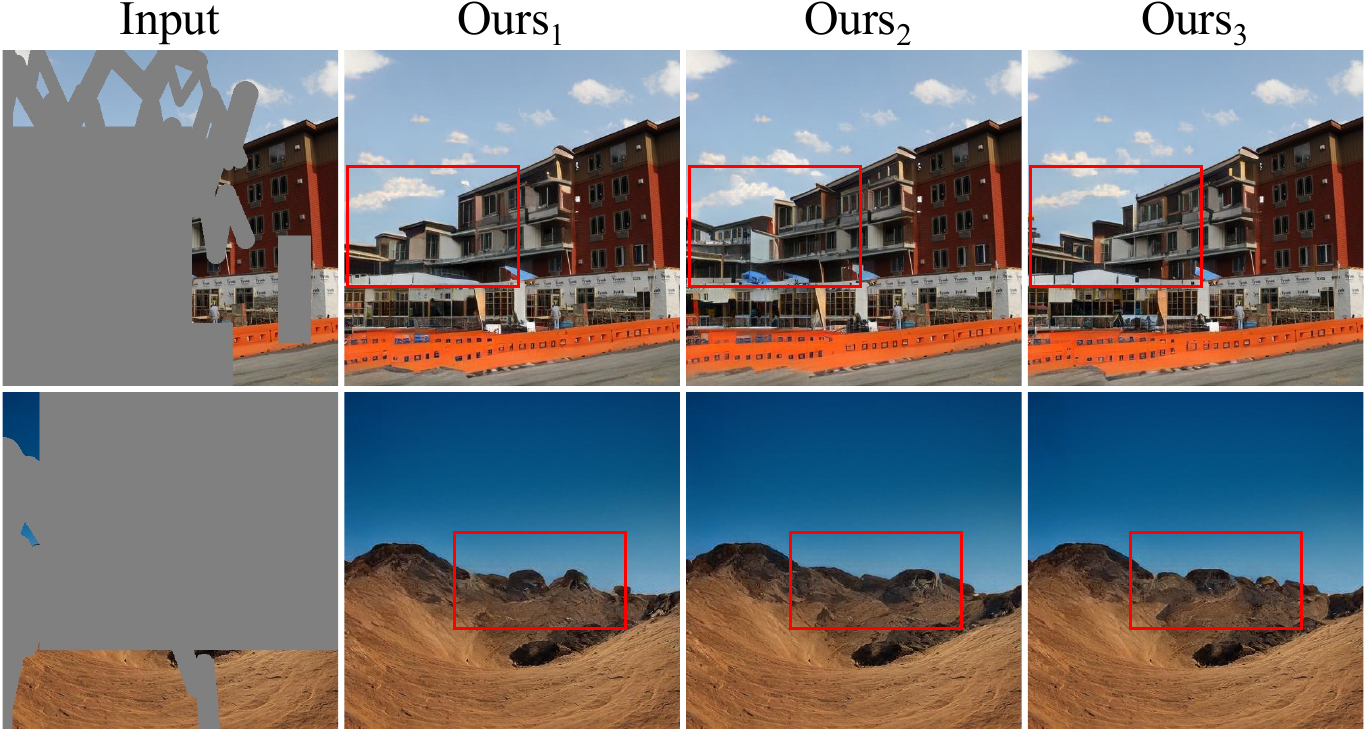}
	\end{center}
	\vspace{-0.05in}
	\caption{Visual examples of diverse generation for our method. The differences mainly lie in the fine-grained details.}
	\vspace{-0.1in}
	\label{fig:diversity}
\end{figure}

\vspace{0.05in}
\noindent\textbf{Continuous sampling.} Our approach may use the estimated variance to perform continuous sampling. Table~\ref{tab:ablation} indicates that FID decreases by nearly 0.1 when continuous sampling (model "E") is not involved. Also, it is observed that our full model leads to more visually consistent content. For example, box structures are well restored according to the visible pixels in Fig.~\ref{fig:abl_vis}. Thus, continuous sampling brings higher fidelity to our results. As shown in Fig.~\ref{fig:diversity}, our model also supports the pluralistic generation, particularly in the hole's center.  However, when the mean term is estimated with low uncertainty or the iteration number is constrained, the differences in results are not always instantly obvious. A detailed analysis of fidelity-diversity trade-off is further provided in the supplementary file.

\vspace{0.05in}
\noindent\textbf{Uncertainty-guided attention.} To fully exploit distant context, we add attention blocks to our framework. We first compare using attention at $32 \times 32$, $16 \times 16$ (model ``A'') and only at $16 \times 16$ (model ``G''). We discover a 0.28 FID drop in model ``G'' from the quantitative comparison in Table~\ref{tab:ablation}, demonstrating the significance of long-range interaction in large-hole image inpainting. 

Besides, as aforementioned in Sec.~\ref{sec:arch}, the conventional attention mechanism may result in color consistency and blurriness. To support this claim, we tease apart the uncertainty guidance and notice a minor performance drop in Table~\ref{tab:ablation}. Also, we provide a visual comparison in Fig.~\ref{fig:abl_vis}. We observe that model ``A'' produces more visually appealing window details than model ``F''.

\vspace{0.05in}
\noindent\textbf{Mask schedule.} As illustrated in Table~\ref{tab:sche_tab} and Fig.~\ref{fig:sche_vis}, we analyze various mask schedule strategies and discover that the uniform strategy achieves the best FID. We argue that this is because the mask ratios of input images vary widely, and uniform schedule results in more stable training for different iterations.

\begin{table}[t]
	\begin{minipage}[c]{0.5\linewidth}
		\centering
		\renewcommand\arraystretch{1.3}
		\resizebox{0.99\linewidth}{!}{
			\begin{tabular}{l c c}
				\toprule[1pt]
				Mask Sche. & Iter. & FID$\downarrow$ \\
				\hline
				Cubic          & 3 & 2.54 \\
				Cosine         & 3 & 2.48 \\
				Linear          & 3 & \textbf{2.36}\\
				Square Root & 3 & 2.47 \\
				\bottomrule[1pt]
			\end{tabular}	
		}
		\vspace{0.02in}
		\caption{Ablation study of mask schedule functions.}
		\label{tab:sche_tab}
	\end{minipage}\hfill
	\begin{minipage}[c]{0.47\linewidth}
		\centering
		\includegraphics[width=1.0\textwidth]{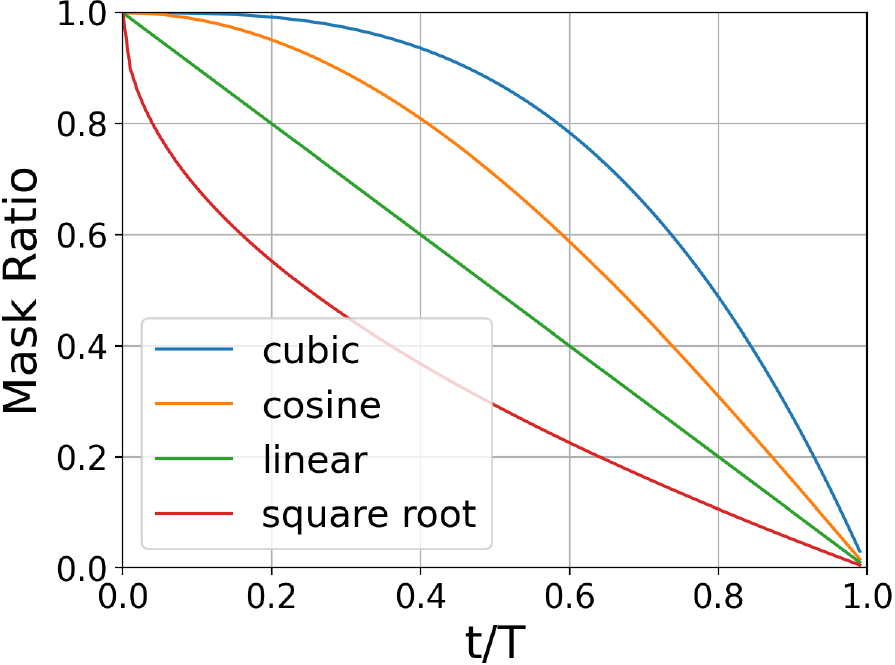}
		\vspace{-0.2in}
		\captionof{figure}{Visualization of mask schedule functions.}
		\label{fig:sche_vis}
	\end{minipage}
\end{table}


\begin{figure*}[t]
	\begin{center}
		\includegraphics[width=1.0\linewidth]{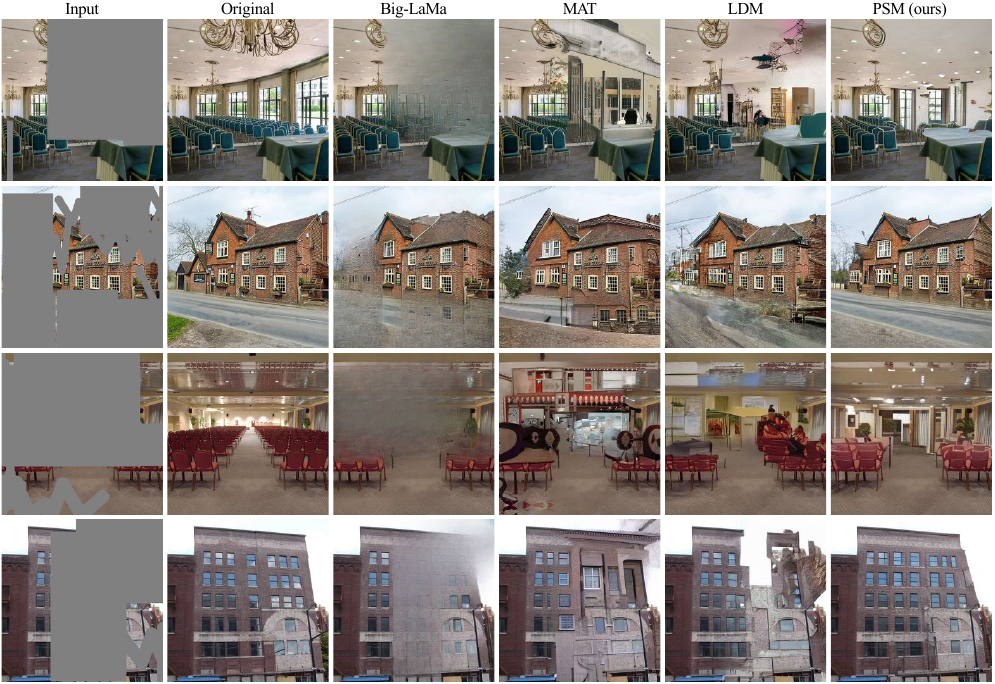}
	\end{center}
	\vspace{-0.1in}
	\caption{Qualitative side-by-side comparisons of state-of-the-art methods on $512\times512$ Places2. Our PSM produces structures and details that are more realistic and reasonable.}
	\vspace{-0.1in}
	\label{fig:sota}
\end{figure*}

\subsection{Comparisons to State-of-the-Art Methods}

We thoroughly compare the proposed pixel spread model (PSM) with GANs-based models~\cite{li2022mat,zhao2020large,suvorov2021resolution,zhu2021image,zeng2021aggregated,yi2020contextual,yu2019free}, autoregressive models~\cite{wan2021high}, and denoising diffusion models~\cite{rombach2022high} in Table~\ref{tab:sota}. Although the last two lines have lately made notable progress even for commercial use, the majority of off-the-shelf techniques can only handle images with up to $256 \times 256$ resolution. We use publicly accessible models for $512 \times 512$ resolution and test them on the same masks to make a fair comparison.

As shown in Table~\ref{tab:sota}, our PSM achieves the state-of-the-art performance on Places2 and CelebA-HQ benchmarks under both large and small mask settings. It can be seen that our method significantly performs better than the existing GANs-based models. Besides, even with only $20\%$ of the parameters of strong denoising diffusion model LDM~\cite{rombach2022high}, our method delivers superior results in terms of all metrics. For example, on the Places2 benchmark, our PSM brings about 1.1 improvement on FID and larger gains on P-IDS and U-ID under the large mask setup. As for the inference speed, our PSM costs nearly 250ms to obtain a $512 \times 512$ image, which is $10 \times$ faster than LDM ($\sim$3s). Notably, our model is trained using far fewer samples (our 20M images vs. CoModGAN's~\cite{zhao2020large} 50M images). We observe that more extended training can further boost performance. All these results demonstrate the effectiveness of our method.

We also provide visual comparisons in Fig.~\ref{fig:sota}. In a variety of scenes, our method generates more aesthetically pleasing textures with fewer artifacts when compared to existing methods. For instance, room layouts and building structures are better inpainted by our approach. More examples are provided in the supplementary materials.


\section{Conclusion}
\label{sec:con}

We have proposed a new pixel spread model for large-hole image inpainting. Utilizing the proposed iteratively decoupled probabilistic modeling, our method can assess the prediction accuracy and retain the pixels with the lowest uncertainty as hints for subsequent processing, yielding high-quality completion. Furthermore, our approach exhibits favorable inference efficiency, significantly surpassing that of prevalent denoising diffusion models. The state-of-the-art performance on multiple benchmarks demonstrate the effectiveness of our method. Additionally, our model has potential for extension to other tasks, such as text-to-image generation, which we will explore in the future.

\noindent\textbf{Limitation analysis.} Our method shows a tendency to make more changes in small details rather than in large structures. We aim to improve the diversity of our generation in this regard. Additionally, our method sometimes struggles to understand objects when only a few hints are given, as illustrated by a few failure cases presented in the supplemental materials.

{\small
\bibliographystyle{ieee_fullname}
\bibliography{egbib}
}


\clearpage

\renewcommand\thesection{\Alph{section}}
\renewcommand\thesubsection{\thesection.\arabic{subsection}}
\renewcommand\thefigure{\Alph{section}.\arabic{figure}}
\renewcommand\thetable{\Alph{section}.\arabic{table}} 

\setcounter{section}{0}
\setcounter{figure}{0}
\setcounter{table}{0}

\twocolumn[
\begin{@twocolumnfalse}
	\begin{center}
		\noindent{\Large{\textbf{Image Inpainting via Iteratively Decoupled Probabilistic Modeling \\ 
					\vspace{0.1in}
					(Supplementary Material)}}}
	\end{center}
	\vspace{0.4in}
\end{@twocolumnfalse}
]

\section{Architecture Details}
\label{sec:arch}

Apart from the descriptions in Sec.~\textcolor{red}{3.2} and Sec.~\textcolor{red}{4.2}, we here provide a more through illustration of architecture details. We adopt a U-Net architecture with skip connections, where the encoder downsamples the size of an $H \times W$ input to $\frac{H}{32} \times \frac{W}{32}$ and the decoder upsamples it back to $H \times W$. At each resolution, there is just one residual block made up of two $3 \times 3$ convolutional layers, unless otherwise stated. Both the encoder and the decoder employ attention blocks at feature sizes of $\frac{H}{16} \times \frac{W}{16}$ and $\frac{H}{32} \times \frac{W}{32}$, and an early convolutional block is also introduced at these scales. Different attention blocks use adaptive mapping functions in Fig.~\textcolor{red}{4}, each of which is composed of 4 convolutional layers with a kernel size of $3 \times 3$.

The input consists of 7 channels: 3 for color images, 1 for the initial mask, 1 for the updated mask, 1 for the uncertainty map, and 1 for the time step. The number of channels is initially converted to 64, then doubled after each downsampling, up to a maximum of 512, and the decoder employs a symmetrical setup. The output contains 6 channels: 3 for the mean term and 3 for the log variance term. 

We use a weight modulation technique, where the style representation is derived from an image global feature and a random latent code. As for the global feature, we employ convolutional layers to further downsample the feature size from $\frac{H}{32} \times \frac{W}{32}$ to $\frac{H}{256} \times \frac{W}{256}$ and a global pooling layer to obtain 1$d$ representation. The random latent code is mapped from Gaussian noise by 8 fully connected layers.

\section{Generalization to 1024$\times$1024 Resolution}
\label{sec:celeba}

To evaluate the generalization ability of models, we compare our pixel spread model (PSM), MAT~\cite{li2022mat} and LaMa~\cite{suvorov2021resolution} trained on $512 \times 512$ Places2~\cite{zhou2017places} at the $1024 \times 1024$ resolution. As illustrated in Table~\ref{tab:high}, our PSM performs significantly better than MAT and LaMa on all metrics, despite using fewer training samples. Remarkably, our approach results in an approximately 1.9 FID improvement. We do not involve denoising diffusion models (\eg, LDM) and other GANs-based models (\eg, CoModGAN) for comparisons because scaling them up to the $1024 \times 1024$ resolution is impractical.

\begin{table}[t]
	\renewcommand\arraystretch{1.2}
	\setlength\tabcolsep{3pt}
	\begin{center}
		\resizebox{0.7\linewidth}{!}{
			\begin{tabular}{c | c c c}
				\hline
				Method & FID$\downarrow$ & P-IDS($\%$)$\uparrow$ & U-IDS($\%$)$\uparrow$ \\
				\hline
				PSM (Ours) & \textcolor{red}{3.95} &  \textcolor{red}{14.40} &  \textcolor{red}{32.23} \\
				\hline
				MAT~\cite{li2022mat} & \textcolor{blue}{5.83} & \textcolor{blue}{9.51} & \textcolor{blue}{28.02} \\
				LaMa~\cite{suvorov2021resolution} & 6.31 & 4.98 & 23.24 \\
				\hline  
			\end{tabular}
		}
	\end{center}
	\caption{Quantitative comparisons on $1024 \times 1024$ Places2~\cite{zhou2017places} dataset under the large mask setup by transferring models trained at the $512 \times 512$ resolution. The \textcolor{red}{best} and \textcolor{blue}{second best} results are in red and blue. Our PSM generalizes well to higher resolutions.}
	\label{tab:high}
\end{table}

\section{512$\times$512 LPIPS Results}
\label{sec:lpips}

\begin{table}[t]
	\renewcommand\arraystretch{1.1}
	\begin{center}
		\resizebox{\linewidth}{!}{
			\begin{tabular}{c | c | c c | c c }
				\hline
				\multirow{2}{*}{Method} & \#Param. & \multicolumn{2}{|c}{Places} & \multicolumn{2}{|c}{CelebA-HQ} \\
				\cline{3-6}
				~ & $\times 10^6$ & Small & Large & Small & Large \\
				\hline
				PSM (Ours)$^\dagger$ & 74 & \textcolor{red}{0.084} & \textcolor{red}{0.161} & \textcolor{red}{0.052} & \textcolor{red}{0.099} \\
				\hline
				Stable Diffusion$^\ddagger$ & 860 & 0.148 & 0.220 & - & - \\
				LDM~\cite{rombach2022high} & 387 & 0.100 & 0.190 & - & - \\
				MAT~\cite{li2022mat} & 62 & 0.099 & 0.189 & \textcolor{blue}{0.065} & \textcolor{blue}{0.125} \\
				CoModGAN~\cite{zhao2020large} & 109 & 0.101 & 0.192 & 0.073 & 0.140 \\
				LaMa~\cite{suvorov2021resolution} & 51/27 & \textcolor{blue}{0.086} & \textcolor{blue}{0.166} & 0.075 & 0.143 \\
				ICT~\cite{wan2021high} & 150 & - & - & 0.105 & 0.195 \\
				MADF~\cite{zhu2021image} & 85 & 0.095 & 0.181 & 0.068 & 0.130 \\
				AOT GAN~\cite{zeng2021aggregated} & 15 & 0.101 & 0.195 & 0.074 & 0.145  \\
				HFill~\cite{yi2020contextual} & 3 & 0.148 & 0.284 & - & - \\
				DeepFill v2~\cite{yu2019free} & 4 & 0.113 & 0.213 & 0.117 & 0.221 \\
				EdgeConnect~\cite{nazeri2019edgeconnect} & 22 & 0.114 & 0.275 & 0.101 & 0.208  \\
				\hline  
			\end{tabular}
		}
	\end{center}
	\caption{LPIPS$\downarrow$ results on $512 \times 512$ Places2~\cite{zhou2017places} and CelebA-HQ~\cite{karras2018progressive} datasets. ``$\dagger$'': our models are trained with 20M samples, much less than other methods (\eg, MAT uses 50M samples on Places2 and 25M samples on CelebA-HQ). We use a single model for both the small and large mask setups. ``$\ddagger$'': the official Stable Diffusion inpainting model is trained on a large-scale high-quality dataset LAION-Aesthetics V2 5+. The \textcolor{red}{best} and \textcolor{blue}{second best} results are in red and blue.}
	\label{tab:lpips}
\end{table}

LPIPS~\cite{zhang2018unreasonable} is also a widely used perceptual metric in image inpainting. For a comprehensive comparison with state-of-the-art methods, we provide LPIPS results in Table~\ref{tab:lpips}. We argue that LPIPS may not be suitable for large-hole image inpainting because it is calculated pixel-by-pixel. This measure is for reference only.

\section{256$\times$256 CelebA-HQ Results}
\label{sec:celeba}

We also conduct quantitative comparisons on $256 \times 256$ CelebA-HQ~\cite{karras2018progressive} dataset. As shown in Table~\ref{tab:celeba}, our method achieves the best performance among all methods.

\begin{table}[t]
	\renewcommand\arraystretch{1.2}
	\setlength\tabcolsep{3pt}
	\begin{center}
		\resizebox{\linewidth}{!}{
			\begin{tabular}{c | c c c | c c c}
				\hline
				\multirow{2}{*}{Method} & \multicolumn{3}{|c}{Small Mask} & \multicolumn{3}{|c}{Large Mask} \\
				\cline{2-7}
				~ & FID$\downarrow$ & P-IDS$\uparrow$ & U-IDS$\uparrow$ & FID$\downarrow$ & P-IDS$\uparrow$ & U-IDS$\uparrow$ \\
				\hline
				PSM (Ours)$^\dagger$ & \textcolor{red}{2.58} & \textcolor{red}{21.35} & \textcolor{red}{33.70} & \textcolor{red}{4.57} & \textcolor{red}{14.07} & \textcolor{red}{25.28} \\
				\hline
				MAT~\cite{li2022mat} & \textcolor{blue}{2.94} & \textcolor{blue}{20.88} & \textcolor{blue}{32.01} & \textcolor{blue}{5.16} & \textcolor{blue}{13.90} & \textcolor{blue}{25.13} \\
				LaMa~\cite{suvorov2021resolution} & 3.98 & 8.82 & 22.57  & 8.75 & 2.34 & 8.77 \\
				ICT~\cite{wan2021high} & 5.24 & 4.51 & 17.39 & 10.92 & 0.90 & 5.23 \\
				MADF~\cite{zhu2021image} & 10.43 & 6.25 & 14.62 & 23.59 & 0.50 & 1.44 \\
				AOT GAN~\cite{zeng2021aggregated} & 9.64 & 5.61 & 14.62 & 22.91 & 0.47 & 1.65 \\
				DeepFill v2~\cite{yu2019free} & 5.69 & 6.62 & 16.82 & 13.23 & 0.84 & 2.62 \\
				EdgeConnect~\cite{nazeri2019edgeconnect} & 5.24 & 5.61 & 15.65 & 12.16 & 0.84 & 2.31 \\
				\hline  
			\end{tabular}
		}
	\end{center}
	\caption{Quantitative comparisons on $256 \times 256$ CelebA-HQ~\cite{karras2018progressive} dataset. The P-IDS and U-IDS results are shown in percentage ($\%$). ``$\dagger$'': our model is trained with 12M samples, far less than other methods (\eg, MAT uses 25M samples). We use a single model for both the small and large mask setups. The \textcolor{red}{best} and \textcolor{blue}{second best} results are in red and blue.}
	\label{tab:celeba}
\end{table}

\section{Comparison to RePaint}
\label{sec:repaint}

Considering that the sizes of RePaint~\cite{lugmayr2022repaint} results are at $256\times 256$ on Places2 and CelebA-HQ while ours are at $512 \times 512$, we don't compare it in the main body of the paper. Here we compare our model PSM to RePaint at $256\times 256$ resolution on Places2 and CelebA-HQ in Table~\ref{tab:repaint}, where PSM achieves better performance and is $1000\times$ faster than RePaint ({\it i.e.}, 0.25s v.s. 250s for one image processing). For saving time, we just use the first 10K Places2 validation images for evaluation. 

\begin{table}[t]
	\begin{center}
		\setlength\tabcolsep{1.6pt}
		\renewcommand\arraystretch{1.2}
		\resizebox{\linewidth}{!}{
			\begin{tabular}{c | c c c | c c c}
				\hline
				\multirow{2}{*}{Method} & \multicolumn{3}{|c|}{Places2-10K ($256 \times 256$)} & \multicolumn{3}{|c}{CelebA-HQ ($256 \times 256$)} \\
				\cline{2-7}
				~ & FID$\downarrow$ & P-IDS($\%$)$\uparrow$ & U-IDS($\%$)$\uparrow$ & FID$\downarrow$& P-IDS($\%$)$\uparrow$ & U-IDS($\%$)$\uparrow$ \\
				\hline
				Ours & \textbf{3.47} & \textbf{18.32} & \textbf{34.52} & \textbf{4.57} & \textbf{14.07} & \textbf{25.28} \\
				RePaint & 6.15 & 11.11 & 27.16 & 10.55 & 0.07 & 1.47 \\
				\hline
			\end{tabular}
		}
	\end{center}
	\caption{Quantitative comparisons with RePaint~\cite{lugmayr2022repaint} on $256 \times 256$ Places~\cite{zhou2017places} and CelebA-HQ~\cite{karras2018progressive} datasets.}
	\label{tab:repaint}
\end{table}

\section{Fidelity-Diversity Trade-Off}
\label{sec:trade}

Apart from FID (depending on both diversity and fidelity), we follow previous work to use Improved Precision and Recall as fidelity (precision) and diversity (recall) measures. As shown in Table~\ref{tab:trade}, our model yields better FID, higher precision yet slightly lower recall than LDM on Places2, while outperforming MAT on all metrics. Improving diversity will be our future work.

\begin{table}[t]
	\begin{center}
		\resizebox{0.9\linewidth}{!}{
			\begin{tabular}{c | c | c c c}
				\hline
				Method & \#Param. & FID$\downarrow$ & Precision$\uparrow$ & Recall$\uparrow$ \\
				\hline
				PSM (Ours) & 74M & \textbf{1.68} & \textbf{0.983} & 0.971 \\
				LDM & 387M & 2.76 & 0.962 & \textbf{0.975} \\
				MAT & 62M & 2.90 & 0.965 & 0.939 \\
				\hline
			\end{tabular}
		}
	\end{center}
	\caption{FID, precision and recall comparisons for evaluating fidelity-diversity traed-off on $512 \times 512$ Places~\cite{zhou2017places} dataset.}
	\label{tab:trade}
\end{table}

\section{Additional Pluralistic Generation}
\label{sec:plura}

\begin{figure}[t]
	\begin{center}
		\includegraphics[width=1.0\linewidth]{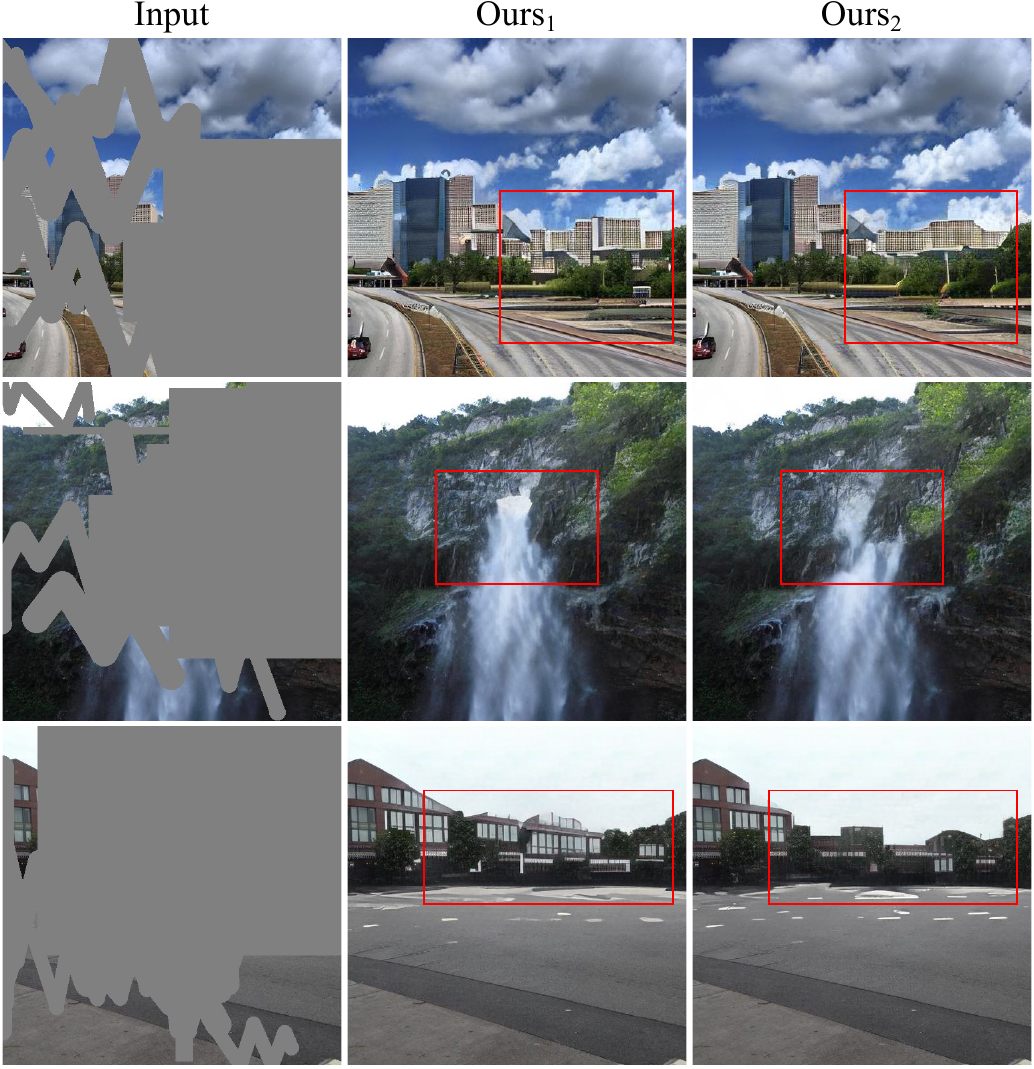}
	\end{center}
	\caption{Visual examples of diverse generation for our method.}
	\label{fig:div}
\end{figure}

As shown in Fig.~\textcolor{red}{6} and Sec.~\textcolor{red}{5}, our method also supports pluralistic generation. From the visual examples in Fig.~\ref{fig:div}, we observe that the differences mainly lie in the fine-grained details. We will work on improving the generating diversity.

\section{Additional Qualitative Comparisons}
\label{sec:quali}

\begin{figure*}[t]
	\begin{center}
		\includegraphics[width=1.0\linewidth]{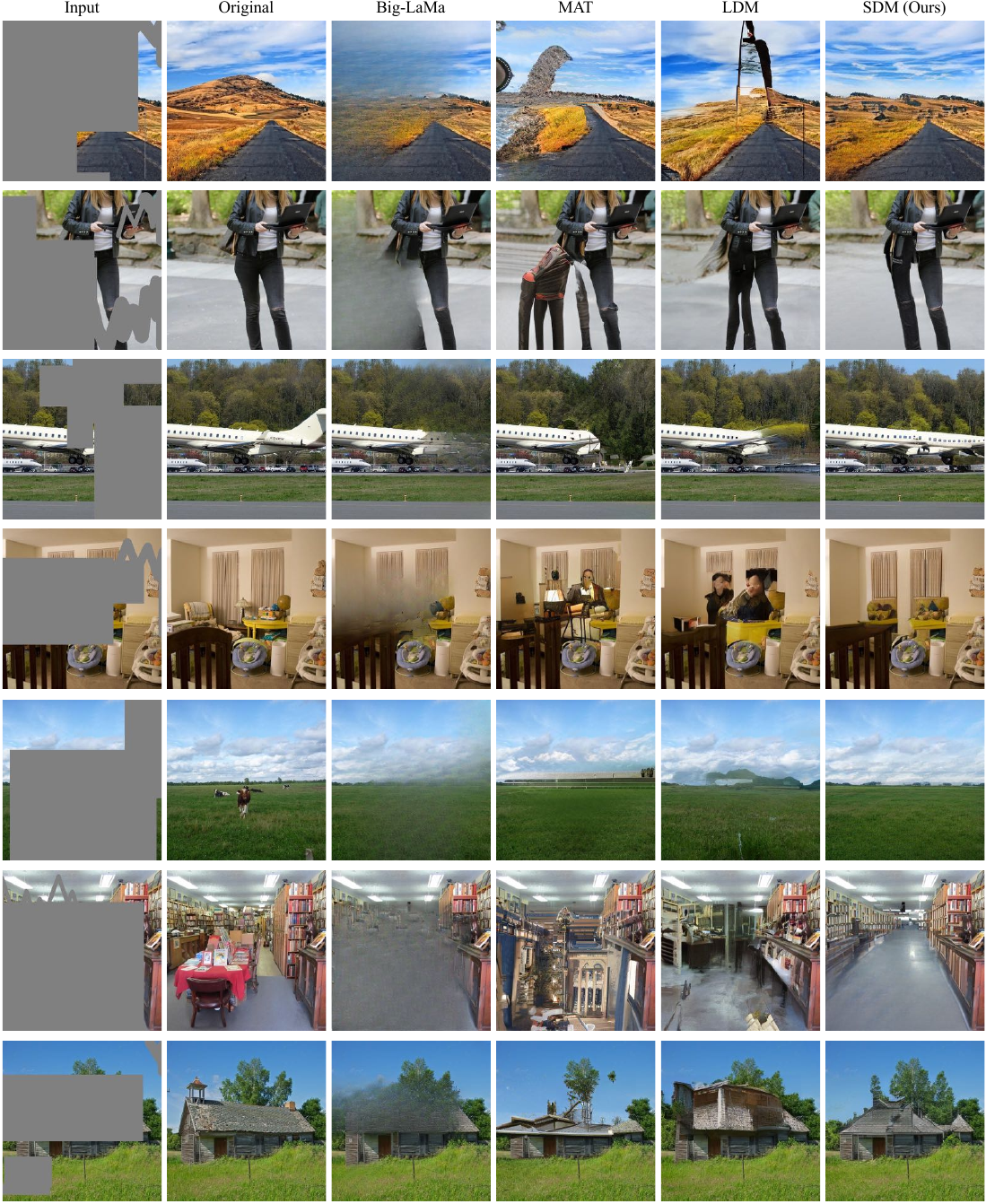}
	\end{center}
	\caption{Qualitative side-by-side comparisons of state-of-the-art methods on $512 \times 512$ Places2 dataset. Please zoom in for a better view. Our PSM produces structures and details that are more realistic and reasonable.}
	\label{fig:sota1}
\end{figure*}

\begin{figure*}[t]
	\begin{center}
		\includegraphics[width=1.0\linewidth]{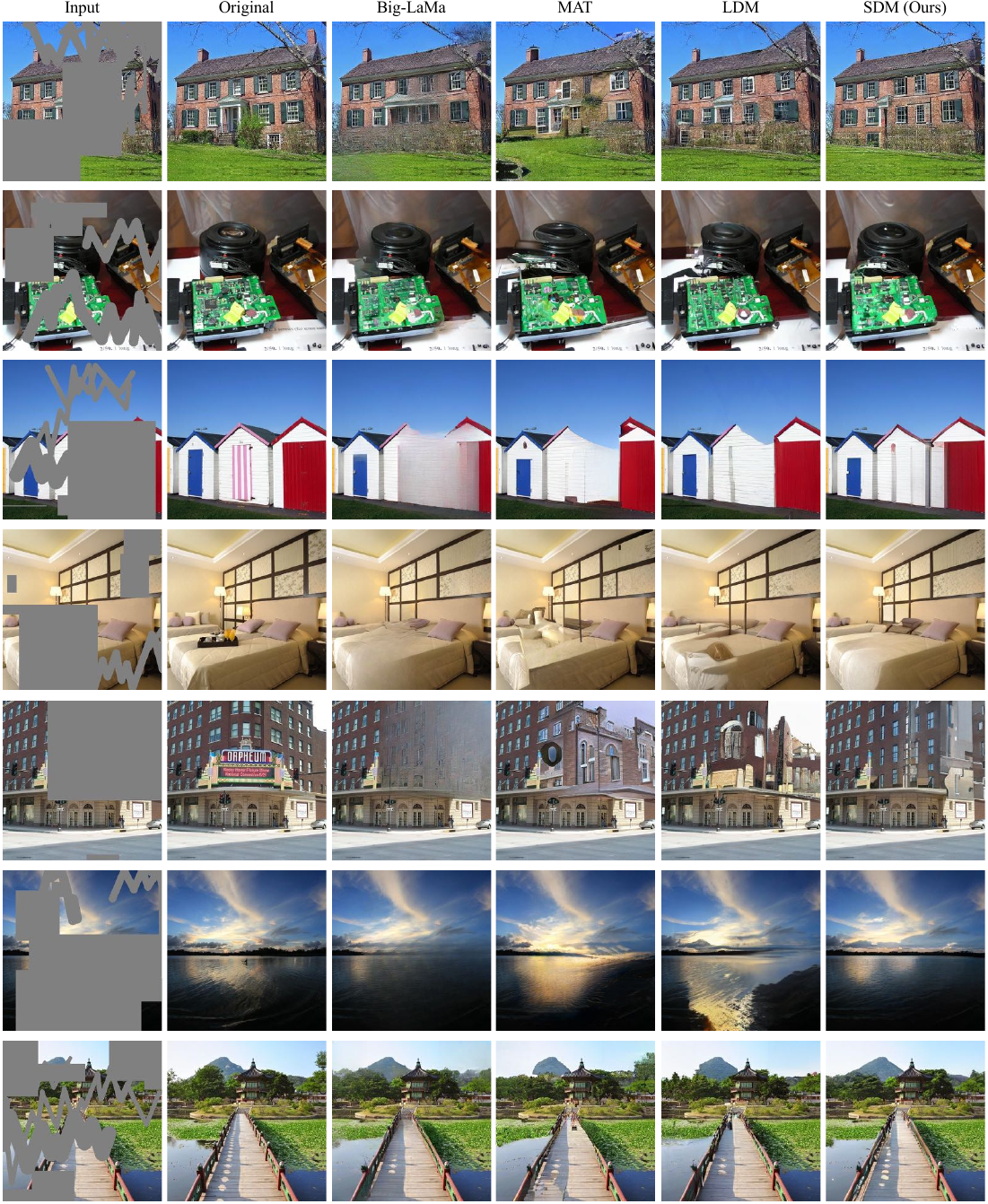}
	\end{center}
	\caption{Qualitative side-by-side comparisons of state-of-the-art methods on $512 \times 512$ Places2 dataset. Please zoom in for a better view. Our PSM produces structures and details that are more realistic and reasonable.}
	\label{fig:sota2}
\end{figure*}

\begin{figure*}[t]
	\begin{center}
		\includegraphics[width=1.0\linewidth]{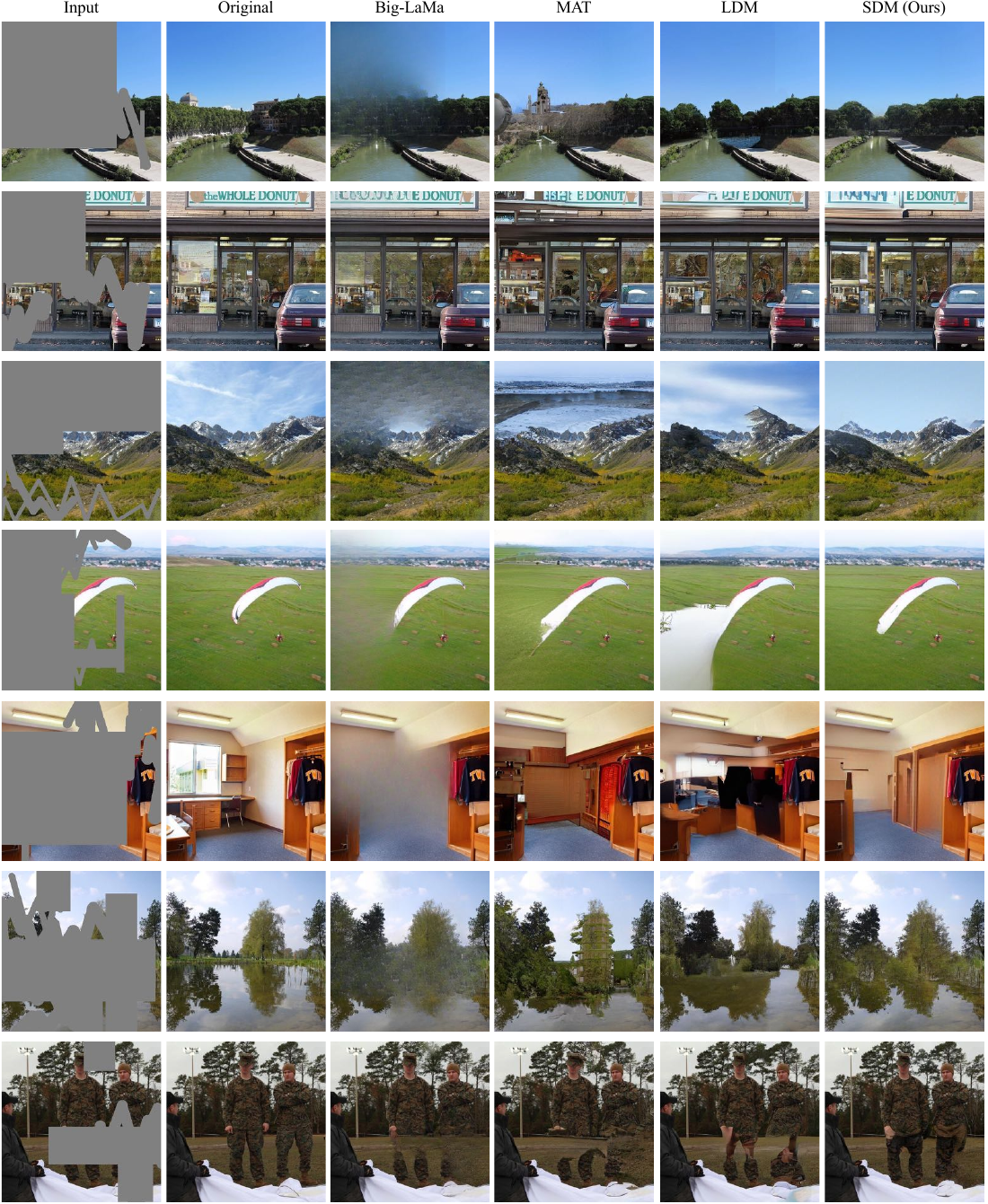}
	\end{center}
	\caption{Qualitative side-by-side comparisons of state-of-the-art methods on $512 \times 512$ Places2 dataset. Please zoom in for a better view. Our PSM produces structures and details that are more realistic and reasonable.}
	\label{fig:sota3}
\end{figure*}

\begin{figure*}[t]
	\begin{center}
		\includegraphics[width=1.0\linewidth]{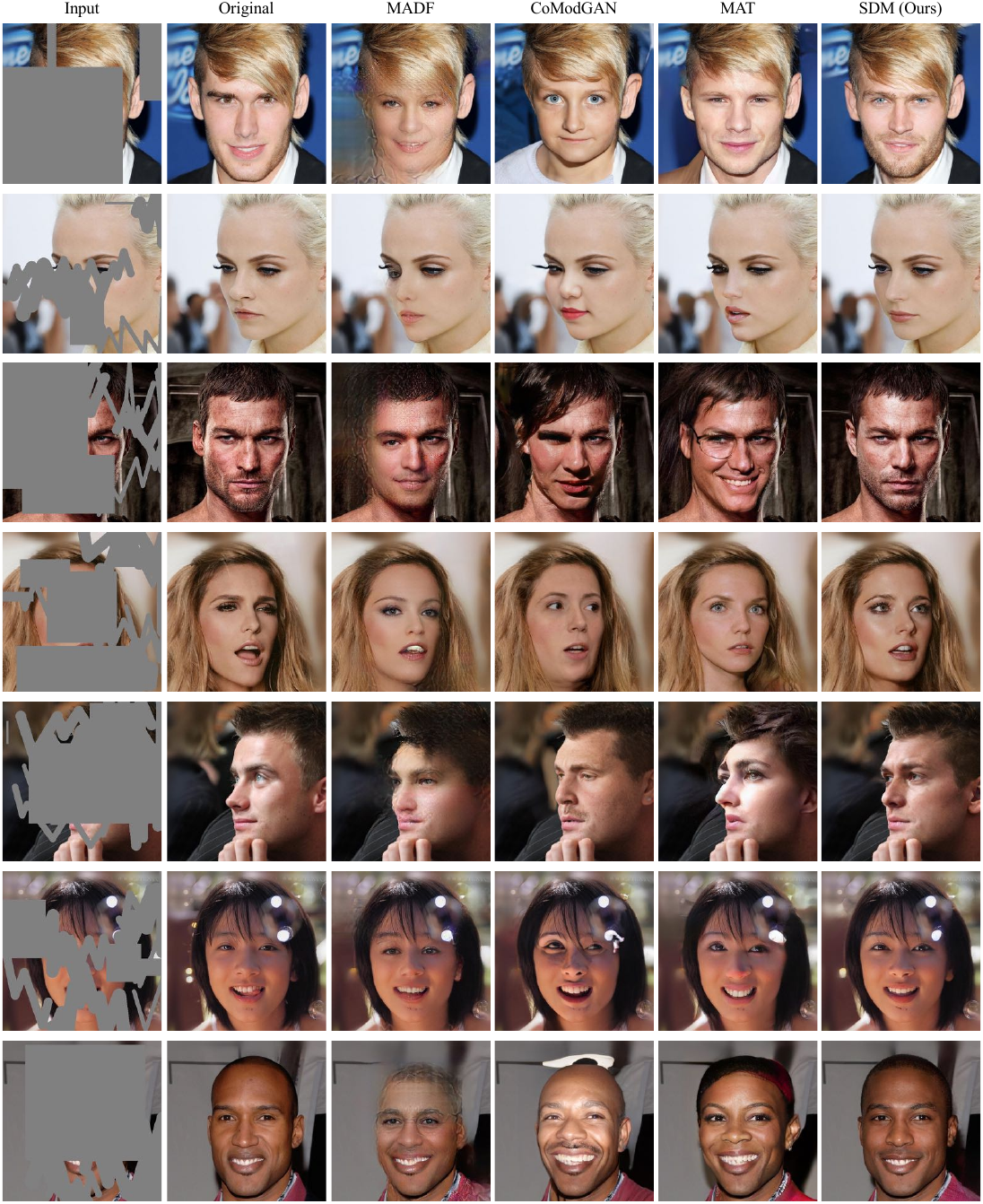}
	\end{center}
	\caption{Qualitative side-by-side comparisons of state-of-the-art methods on $512 \times 512$ CelebA-HQ dataset. Please zoom in for a better view. Our PSM produces face outlines and details that are more realistic and reasonable.}
	\label{fig:sota4}
\end{figure*}

\begin{figure*}[t]
	\begin{center}
		\includegraphics[width=1.0\linewidth]{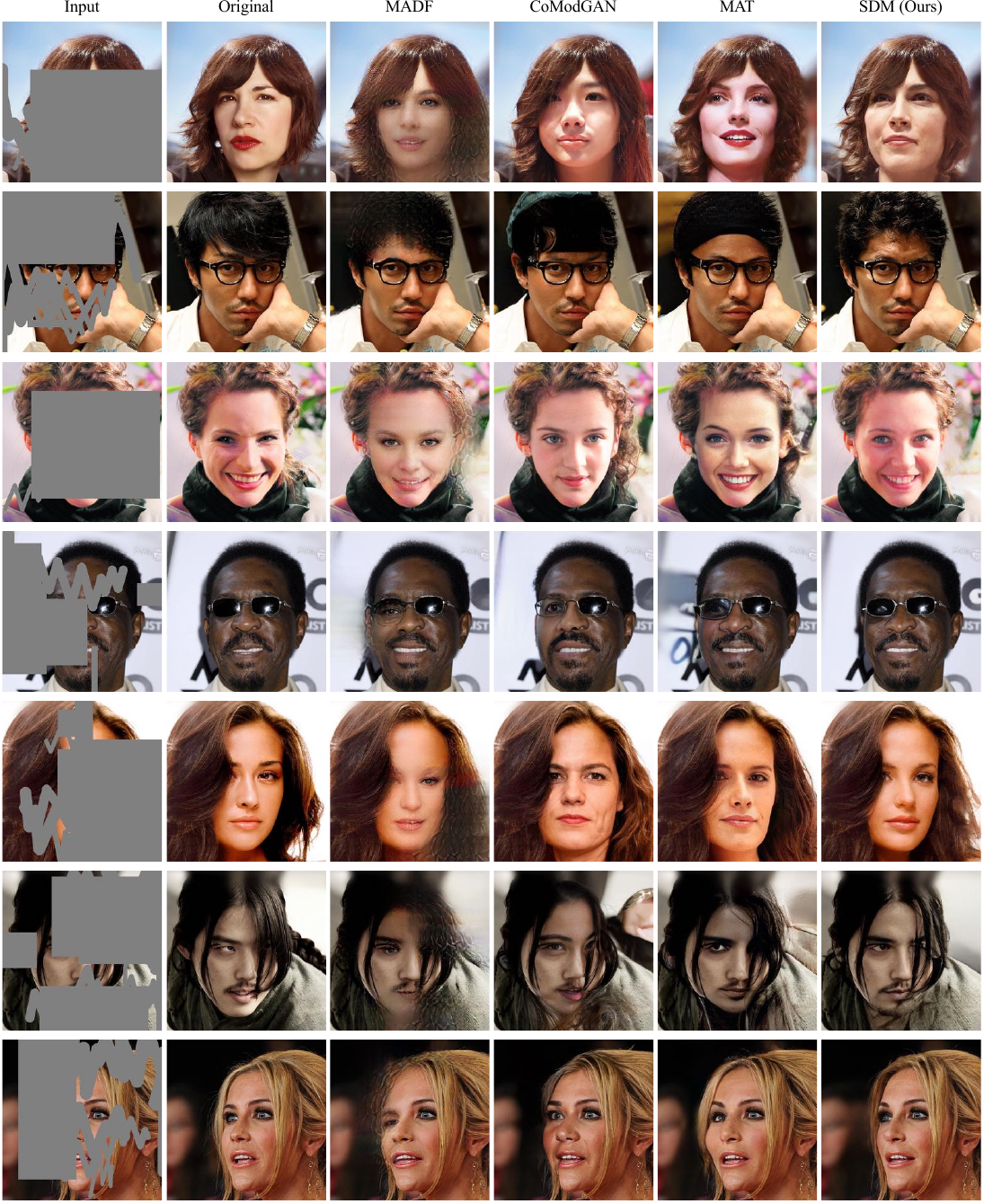}
	\end{center}
	\caption{Qualitative side-by-side comparisons of state-of-the-art methods on $512 \times 512$ CelebA-HQ dataset. Please zoom in for a better view. Our PSM produces face outlines and details that are more realistic and reasonable.}
	\label{fig:sota5}
\end{figure*}

\begin{figure*}[t]
	\begin{center}
		\includegraphics[width=1.0\linewidth]{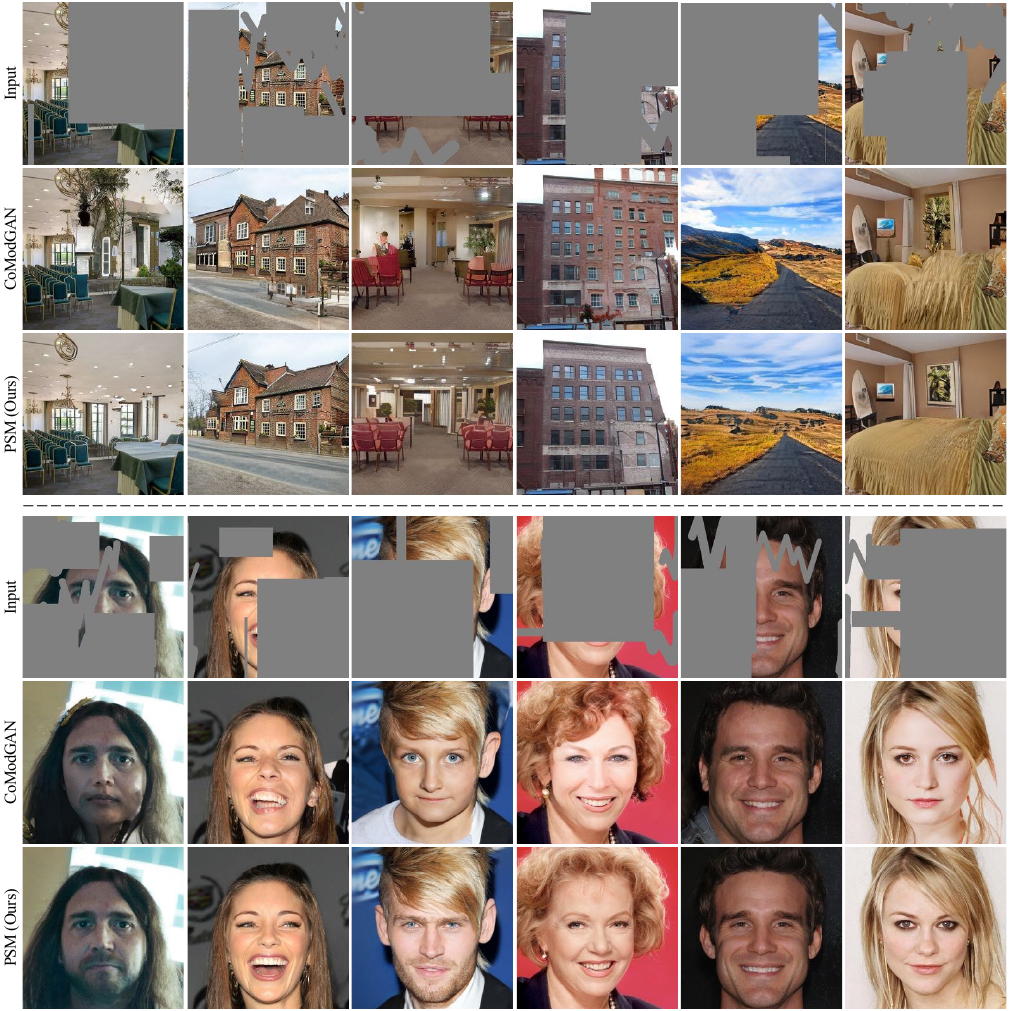}
	\end{center}
	\caption{Qualitative comparisons between CoModGAN and our PSM on $512 \times 512$ Places2 and CelebA-HQ datasets. Please zoom in for a better view. Our PSM produces structures and details that are more realistic and reasonable.}
	\label{fig:comodgan}
\end{figure*}

We provide more visual examples on $512 \times 512$ Places2~\cite{zhou2017places} and CelebA-HQ~\cite{karras2018progressive} in Fig.~\ref{fig:sota1}, Fig.~\ref{fig:sota2}, Fig.~\ref{fig:sota3}, Fig.~\ref{fig:sota4}, and Fig.~\ref{fig:sota5}. Due to space limit, we additionally add comparisons with CoModGAN~\cite{zhao2020large} in Fig.~\ref{fig:comodgan}. Compared to other methods, our method generates more photo-realistic and semantically consistent content. For example, our method successfully recovers human legs, airplane structures, and more realistic indoor and outdoor scenes. All the results demonstrate the effectiveness of our method. 

\section{Failure Cases}
\label{sec:fail}

\begin{figure}[t]
	\begin{center}
		\includegraphics[width=1.0\linewidth]{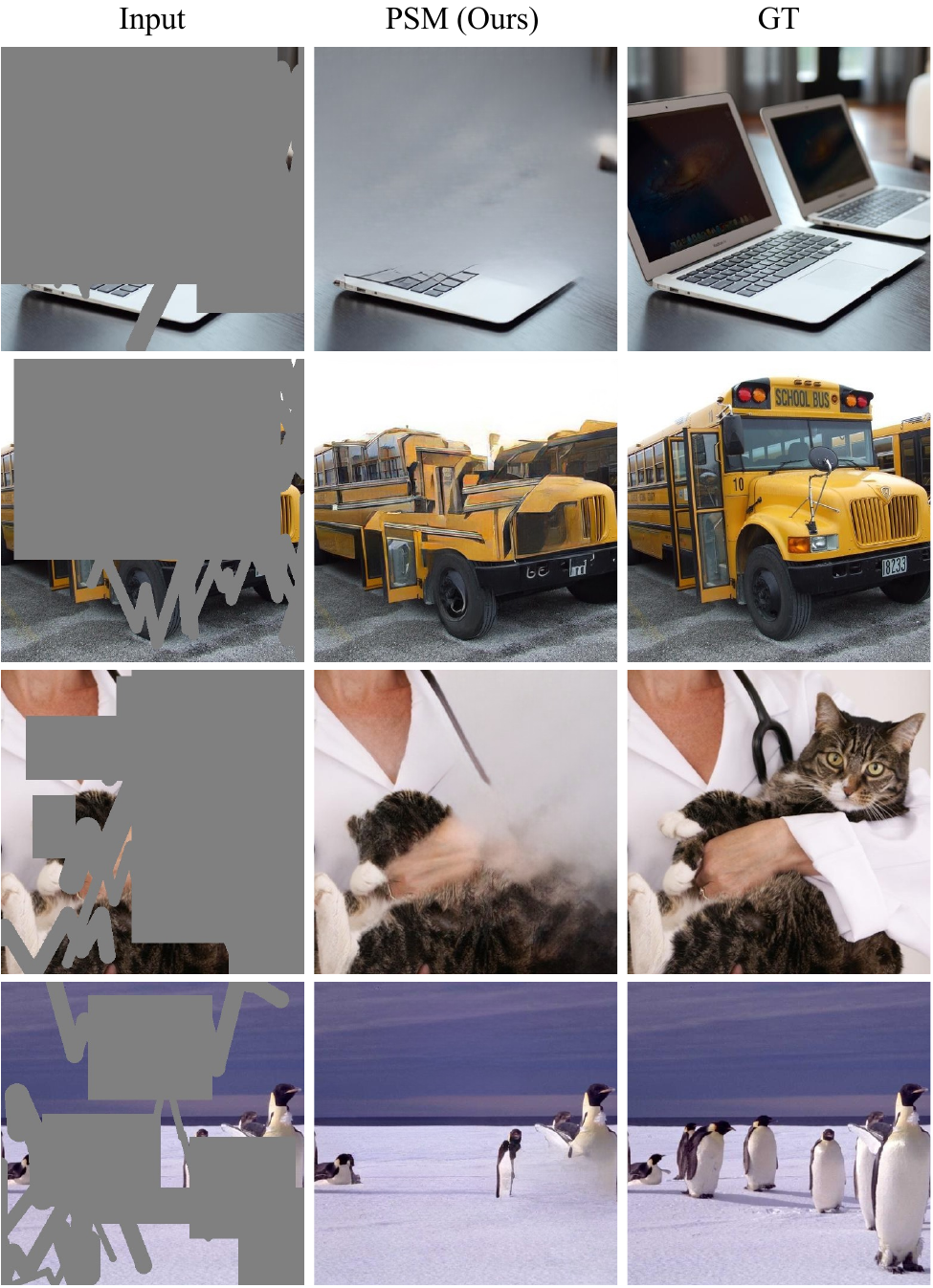}
	\end{center}
	\caption{Failure cases of our PSM. It is difficult to recover the objects when large-scale regions are missing.}
	\label{fig:failure}
\end{figure}

As discussed in Sec.~\textcolor{red}{5}, our model sometimes fails to recover the damaged objects when limited clues are provided. We show some failure cases in Fig.~\ref{fig:failure}. For instance, the missing part of the notebook is filled with the background, and the recovered bus structure is incomplete. We attribute one of the reasons to the lack of high-level semantic understanding. We will further improve the generative capability of our model.


\end{document}